\documentclass{article}

\usepackage{arxiv}

\usepackage[utf8]{inputenc} % allow utf-8 input
\usepackage[T1]{fontenc}    % use 8-bit T1 fonts
\usepackage{hyperref}       % hyperlinks
\usepackage{url}            % simple URL typesetting
\usepackage{booktabs}       % professional-quality tables
\usepackage{amsfonts}       % blackboard math symbols
\usepackage{nicefrac}       % compact symbols for 1/2, etc.
\usepackage{microtype}      % microtypography
\usepackage{lipsum}
\usepackage{graphicx}

\usepackage{graphicx}
\usepackage{amsmath}
\usepackage{amssymb}
\usepackage{booktabs}

\usepackage{subcaption}
\usepackage{afterpage}

\newcommand{\etal}{\emph{et al.}}
\newcommand{\eg}{\emph{e.g.}}

\newtheorem{definition}{Definition}[section]
\newtheorem{theorem}{Theorem}[section]

\graphicspath{ {./Figures/} }

\usepackage[capitalize]{cleveref}
\crefname{section}{Sec.}{Secs.}
\Crefname{section}{section}{sections}
\Crefname{table}{Table}{Tables}
\crefname{table}{Tab.}{Tabs.}

\title{Interpretable Dimensionality Reduction by Feature Preserving Manifold Approximation and Projection}

\author{
 Yang Yang\\
  Frazer Institute\\
  University of Queensland\\
  \texttt{yang.yang1@uq.edu.au} \\
   \And
 Hongjian Sun \\
  Frazer Institute\\
  University of Queensland\\
  \texttt{hongjian.sun@uq.edu.au} \\
  \And
 Jialei Gong \\
  Frazer Institute\\
  University of Queensland\\
  \texttt{j.gong@uq.edu.au} \\
  \And
 Di Yu \\
  Frazer Institute\\
  University of Queensland\\
  \texttt{di.yu@uq.edu.au} \\
}

\begin{document}
\maketitle
%%%%%%%%% ABSTRACT
\begin{abstract}
%   The ABSTRACT is to be in fully justified italicized text, at the top of the left-hand column, below the author and affiliation information.
%   Use the word ``Abstract'' as the title, in 12-point Times, boldface type, centered relative to the column, initially capitalized.
%   The abstract is to be in 10-point, single-spaced type.
%   Leave two blank lines after the Abstract, then begin the main text.
%   Look at previous CVPR abstracts to get a feel for style and length.

% source features
% tangent space -> gauge embedding
% object detection
% explain adversarial results
% Code is avaiable

% lack interpretability due to missing source features

% featMAP preserves source features by tangent space embedding
% local SVD; tangent space alignment; 
% Anisotropic projection

% Maintain both source features and topological structure

Nonlinear dimensionality reduction lacks interpretability due to the absence of source features in low-dimensional embedding space. 
We propose an interpretable method featMAP to preserve source features by tangent space embedding.
The core of our proposal is to utilize local singular value decomposition (SVD) to approximate the tangent space which is embedded to low-dimensional space by maintaining the alignment. 
Based on the embedding tangent space, featMAP enables the interpretability by locally demonstrating the source features and feature importance.
Furthermore, featMAP embeds the data points by anisotropic projection to preserve the local similarity and original density.
We apply featMAP to interpreting digit classification, object detection and MNIST adversarial examples. FeatMAP uses source features to explicitly distinguish the digits and objects and to explain the misclassification of adversarial examples. 
We also compare featMAP with other state-of-the-art methods on local and global metrics.

\end{abstract}

%%%%%%%%% BODY TEXT
%\vspace{-1mm}
\section{Introduction}
%\vspace{-1mm}
\label{sec:intro}

% Intuition for feature importance.
% Locally the directions with the largest variance incorporate majority of feature information, and these directions belong to the tangent space.

\begin{figure}[t]
  %\vspace{-3mm}
  \centering
%   \fbox{\rule{0pt}{2in} \rule{0.9\linewidth}{0pt}}
  \includegraphics[width=0.5\linewidth]{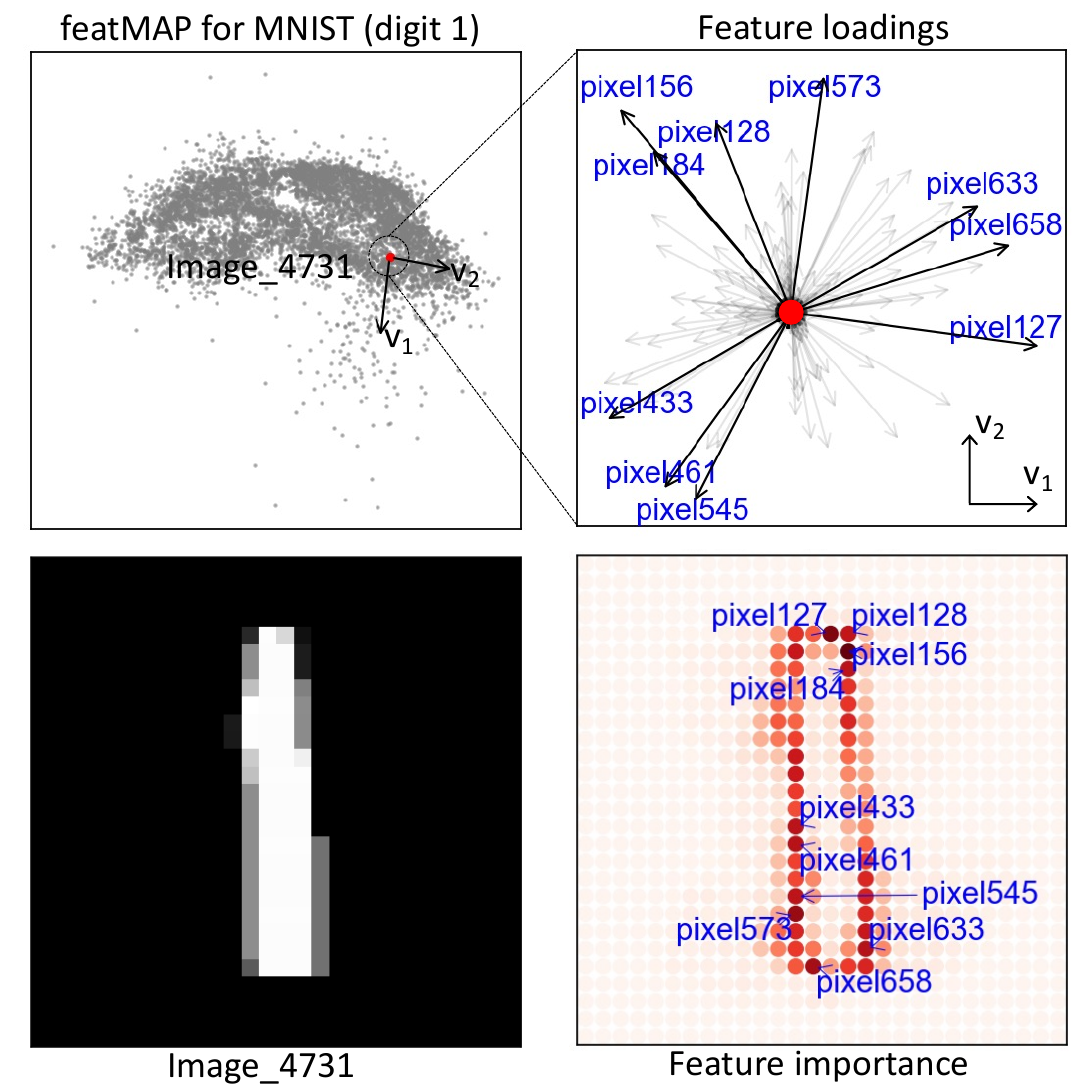}

  \caption{FeatMAP preserving source features.
  FeatMAP embeds the digit $1$ group of MNIST to two-dimensional space (top-left). 
  One randomly selected data point (in red) is associated with the embedding tangent space ($\mathit{span}(v_1, v_2)$) showing the source features (top-$10$ annotated) of every pixel (top-right). The feature importance computed by the feature loadings is mapped to the selected image (bottom-right), detecting the digit edge. }
  \label{fig:feat_load}
  %\vspace{-7mm}
\end{figure}    

Nonlinear dimensionality reduction methods are ubiquitously applied for visualization and preprocessing high-dimensional data in machine learning~\cite{tenenbaum2000global,roweis2000nonlinear,zhang2006mlle,donoho2003hessian,belkin2003laplacian,zhang2004principal,van2008visualizing,mcinnes2018umap}. % More citation like diffusion map
% Based on nearest neighbour graph, 
These methods assume that the intrinsic dimension of the underlying manifold is much lower than the ambient dimension of the real-world data~\cite{levina2004maximum,pope2021intrinsic,wright2022high}.
Based on approximating the manifold by $k$ nearest neighbour ($k$NN) graph, nonlinear dimensionality reduction projects data from high to low-dimensional space and retains the topological structure of original data.
% The reason for applying dimensionality reduction is that the intrinsic dimension of the real-world data is much lower than the ambient dimension~\cite{levina2004maximum,pope2021intrinsic,wright2022high}.
% The low-dimensional projection capturing the structure of original data 
% nonlinear dimensionality reduction is critical to the efficient and effective computation in visualizing the classification~\cite{van2008visualizing,mcinnes2018umap}.
% More for nonlinear DR 

% Weakness
While nonlinear dimensionality reduction is effective for visualizing high-dimensional data, one major weakness is lacking interpretability of the reduced-dimension results~\cite{mcinnes2018umap}.
The reduced dimensions of nonlinear dimensionality reduction have no specific meaning, compared with linear methods like Principal Component Analysis (PCA) where the dimensions of the embedding space represent the directions of the largest variance of original data.
Particularly, nonlinear dimensionality reduction focuses on preserving distance between observations and thereby loses source feature information in the embedding space,
resulting in failing to illustrate feature loadings that linear methods such as PCA can provide to explain the feature contribution in each dimension.

In this paper, we seek to improve the interpretability of nonlinear dimensionality reduction.
% address the problem of nonlinear dimensionality reduction lacking interpretability
% \yali{this sentence seems confusing? maybe we aim to improve interpretability of nonlinear dimensionality reduction? }. 
In addition to preserving the local topological structure between observations in the embedding space, we aim to incorporate the source features to devise an interpretable nonlinear dimensionality reduction method.
The feature information is encoded in the column space of data, and we use the tangent space to locally depict the column space~\cite{singer2012vector,lim2021tangent}.
The intuition of employing tangent space originates from the anisotropic density observation on a manifold such that some curves passing through a point in the manifold are flat while some through the same point are steep, indicating that the feature variation diverges in different directions. 
This equivalence class of tangent curves forms the tangent space which retains source feature information.
% the data density in the manifold is anisotropic
% this anisotropic property is characterized by the tangent space in the manifold~\cite{singer2012vector,lim2021tangent}.

We propose \textbf{feat}ure preserving \textbf{m}anifold \textbf{a}pproximation \textbf{a}nd \textbf{p}rojection (featMAP) to maintain both the pairwise distance on the manifold and the source features on tangent space.
% We first approximate the manifold and tangent space.
We first approximate the manifold topological structure by $k$NN graph~\cite{preparata2012computational,dong2011efficient}
% which is constructed by computing pairwise distances and connecting each data point to its $k$-nearest neighbours~\cite{preparata2012computational,dong2011efficient}.
and compute the tangent space by applying singular value decomposition (SVD) to the local nearest neighbours.
The column space of SVD spans the basis for tangent space~\cite{singer2012vector,lim2021tangent}. 
The tangent spaces of two near data points are connected by parallel transport which is estimated as orthogonal transformation between the bases of tangent space~\cite{singer2012vector}. We refer to the orthogonal transformation as alignment.
% We assign more weight to the local direction with larger singular value.
% What is alignment.
In addition, the tangent space at one data point is associated with a $k$NN point cloud, whose radius is determined by the singular values of SVD.
This local point cloud forms a hyperellipsoid demonstrating the anisotropic density.
For dimensionality reduction, we first embed the tangent space by preserving the alignment between tangent spaces. We achieve this by depicting the alignment with cosine distance and minimizing the difference of cosine distance distribution between high and low-dimensional space.
Based on the embedding tangent space, we project the data points to maintain pairwise distance by minimizing the difference of distance distribution in original and embedding space.  
We also preserve the anisotropic density by maximizing the correlation of local (hyper)ellipsoid radius between high and low-dimensional space.
% Similar to~\cite{van2008visualizing,mcinnes2018umap}, we minimize the distribution difference of high and low-dimensional space to achieve the embedding results.

The embedding tangent space by featMAP provides a frame to illustrate the source features.
% The reduced low-dimensional space by featMAP maintains the topological structure as well as source feature information. 
\Cref{fig:feat_load} depicts an example of embedding MNIST (digit $1$ group) dataset. The embedding coordinate of a randomly selected data point is associated with a local frame showing the source features and feature importance. 
% We map the feature importance back to the corresponding image, and find that the pattern by feature importance explicitly detect the object edge. 
% To summarize, our main contribution is designing an interpretable nonlinear dimensionality reduction method which preserves the source features as well as topological structure in low-dimensional space.
% This enables the interpretability of embedding results such as visualization.
To summarize, we make the following contributions:
% \yali{usually, one contribution would be 'We evaluate feaMap on xxx task. empirical results on xxx tasks show that our method achieves xxx performance.'}

\begin{itemize}
%\vspace{-1.5mm}
\item We propose featMAP, an interpretable nonlinear dimensionality reduction method that preserves source features and local similarity.
% \item We introduce tangent space embedding by maintaining the alignment between tangent spaces in low-dimensional space and provide a frame to locally illustrate features.
% \item We present anisotropic projection which not only preserves the local similarity, but also retains the original density in low-dimensional space.
%\vspace{-1.5mm}
\item We evaluate featMAP on digit and object data. FeatMAP utilizes feature information to successfully explain the digit classification and object detection.
%\vspace{-1.5mm}
\item We apply featMAP to MNIST adversarial examples to explicitly show that the adversarial attack changes the feature importance, which fools the LeNet classifier.
%\vspace{-2mm}
\end{itemize}
In the following sections, we first discuss the related works before delving into the proposed method, followed by experiments on interpreting digit classification, object detection and MNIST adversarial examples, and comparing with the state-of-the-art methods on local and global metrics.

% With the tangent space of any data point incorporating principal directions, the low-dimensional embedding space can also reveal the anisotropic density of original high-dimensional data. Meanwhile, the tangent space also retains the variance of pairwise distance by SVD in local neighbourhood area, thus rectifying the distortion issue. Therefore, the proposed algorithm will improve the interpretability of UMAP in processing high-dimensional single-cell data.

% For any data point, the manifold can be approximated by the tangent space equipped with the inner product, which is the Riemannian manifold.
% For the tangent space of a given data point, we compute the basis vectors which are arranged in descending order of the singular values. A large singular value means that there is large data variance along this basis vector; the feature contribution to basis vectors can be induced by the inner product.

% Figure 1

%\vspace{-2mm}
\section{Related Work}
%\vspace{-1mm}
\label{sec:relatedwork}

% Nonlinear dimensionality reduction by manifold learning achieves high accuracy in distance and neighbourhood preservation~\cite{borg2005modern,van2008visualizing,mcinnes2018umap}, 
The interpretability of nonlinear dimensionality reduction is missing to design and evaluate the embedding methods~\cite{liu2016visualizing,vellido2012making,frenay2016information,dumas2018interaction}. 
Linear methods like PCA naturally possess the interpretability by explicitly showing source features~\cite{gabriel1971biplot}.
Limited research has been conducted to address the interpretability of nonlinear dimensionality reduction.
Liu \etal~\cite{liu2016visualizing} proposed to understand the link of embedding dimensions and the original ones as a trade-off between interpretability and the intrinsic structure of the embedding.
Bibai and Fr\'enay~\cite{bibal2019measuring} introduced to solve the interpretability problem by explaining the low-dimensional axes~\cite{bibal2018finding,marion2019bir} and analysing the data points position in scatter plot~\cite{sips2009selecting}; they further pointed out that the dimensionality reduction is interpretable if the embedding data points can be understood by high-dimensional features.
Recently, Wu \etal~\cite{wu2019solving} applied subspace projection to solving interpretable kernel dimensionality reduction. Bibal \etal~\cite{bibal2020explaining} adapted local interpretable model-agnostic explanations to locally explain t-SNE. Bardos \etal~\cite{bardos2022local} introduced a model-agnostic technique for local explanation of dimensionality reduction.
Our approach aims to incorporate source features in embedding to locally illustrate feature importance and interpret the reduced-dimension results.

Nonlinear dimensionality reduction is claimed to better preserve distance and neighbourhood information in the projected manifold~\cite{borg2005modern,van2008visualizing,mcinnes2018umap}. 
The current state-of-the-art manifold learning methods such as t-SNE~\cite{van2008visualizing}, LargeVIs~\cite{tang2016visualizing} and UMAP~\cite{mcinnes2018umap} create the low-dimensional embedding 
related to the SNE framework~\cite{hinton2002stochastic,bohm2022attraction,damrich2022contrastive}
% by minimizing the difference of distribution (induced by Euclidean distance) in high and low-dimensional space. 
% These methods aim at improving efficiency and preservation of both global and local structure, 
which relies on constructing and embedding $k$NN graph to approximate the underlying manifold. 
% A wide range of dimensionality reduction methods lie within the manifold learning context, including Isomap~\cite{tenenbaum2000global}, Locally Linear Embedding (LLE)~\cite{roweis2000nonlinear}, Modified LLE~\cite{zhang2006mlle}, Hessian Eigenmapping~\cite{donoho2003hessian}, Spectral Embedding~\cite{belkin2003laplacian}, Local Tangent Space Alignment (LTSA)~\cite{zhang2004principal}, Multi-dimensional Scaling (MDS)~\cite{borg2005modern}, t-SNE~\cite{van2008visualizing} and UMAP~\cite{mcinnes2018umap}.
Recently, several methods have attempted to improve the manifold learning methods based on exploiting topological structures. TriMAP~\cite{amid2019trimap} used 2-simplex structure and constructed triple constraints to improve global accuracy. PaCMAP~\cite{wang2021understanding} considered multi-hop structure to keep both local and global properties. 
DensMAP~\cite{narayan2021assessing} computed the radius of local neighbourhood area to maintain the density of high-dimensional data. 
h-NNE~\cite{sarfraz2022hierarchical} applied a hierarchy built on nearest neighbour graph to preserving multi-level grouping properties of original data.
SpaceMAP~\cite{zu2022spacemap} used space expansion to match the high and low-dimensional space.
CO-SNE~\cite{guo2022co} extended t-SNE from Euclidean space to hyperbolic space.
However, these methods lack interpretability due to the absence of source features in the embedding results.
% Incorporating high-order topological structures can improve UMAP in better local and global accuracy; 
% nevertheless, the source features are absent in embedding space and it is imperative to preserve the features to provide interpretability for manifold learning.

% The source features are in the column space which is locally depicted by tangent space. Zhang and Zha~\cite{zhang2004principal} proposed LTSA to approximate the manifold by locally aligning the tangent space, and the tangent space was estimated by local singular value decomposition (SVD)~\cite{zhang2004principal,singer2012vector,lim2021tangent}.
% Tangent space 

Our approach belongs to the paradigm of manifold learning.
Besides maintaining the manifold topological structure, we preserve the source features in tangent space~\cite{zhang2004principal,singer2012vector} and enable the interpretability of dimensionality reduction by locally understanding source features in embedding tangent space. This provides us with a highly effective and interpretable dimensionality reduction method.

% Lack interpretability

\begin{figure*}[t]
%\vspace{-2mm}
  \centering
%   \fbox{\rule{0pt}{2in} \rule{0.9\linewidth}{0pt}}
  \includegraphics[width=\linewidth]{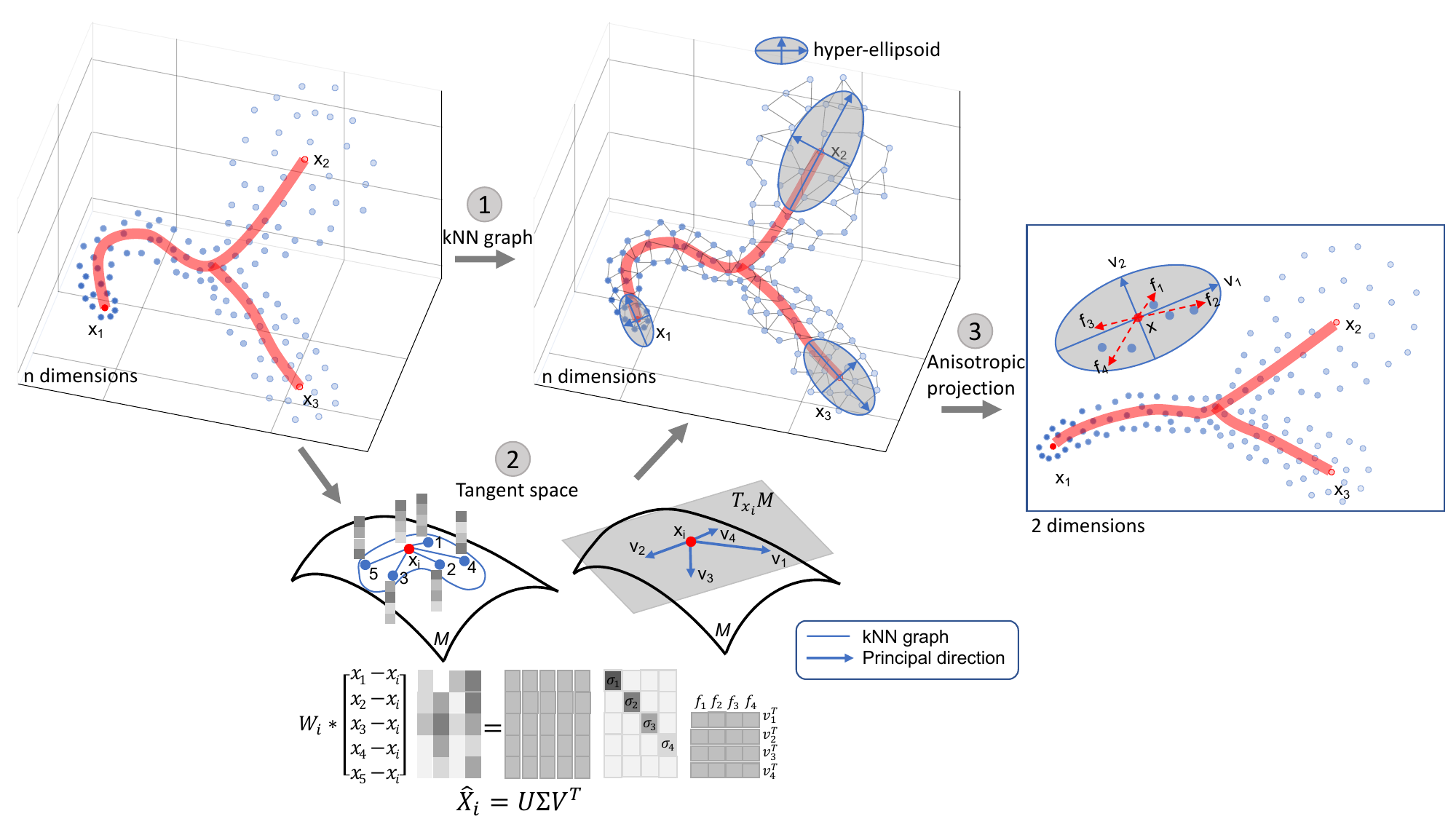}
  \caption{ The framework of featMAP.
  With $n$-dimensional data (top-left) as input, 
  featMAP first constructs the topological space by $k$NN graph (step $1$, top-middle),
  followed by compute the tangent space by local SVD (step $2$, bottom-middle) 
  and embed the tangent space to preserve alignment (\Cref{fig:gauge_emb}).
  Along the embedding tangent space, featMAP applies anisotropic projection (step $3$) to embedding the data to low-dimensional space (right), which locally retains the source features in embedding tangent space.
  Data points with darker blue indicate denser patterns, and vice versa. 
%   \yali{1,2,3 are labeled in the figure, but also need to be cited in caption.}
%   Topological space by knn;
%   Tangent space by local svd;
%   embed tangent space;
%   project data points;
%   The input $n$-dimensional data (top-left) are first 
}
  \label{fig:framework}
%\vspace{-6mm}
\end{figure*}

\section{The featMAP method}
\label{sec:algorithm}
We present featMAP to preserve both manifold structure and source features in low-dimensional space.
%better preserve the topological structure of manifold in high-dimensional space. 
We first characterise the manifold topological structure by $k$NN graph and compute the tangent space by local SVD,
then we embed the tangent space by preserving alignment between tangent spaces,
and project the data points along the embedding tangent space to low-dimensional space. 
In the following sections, we elaborate on each step and validate the proposed methods. \Cref{fig:framework} illustrates the framework of the method.

\subsection{Topological space and feature space approximation}
\label{subsec:tangent_appr}
We follow the manifold assumption that data points $\{x_1,...,x_m\}\footnote{Vectors of lower-case letters refer to column vectors.}\subset \mathbb{R}^n $ lie on a $d$-dimensional Riemannian manifold $\mathcal{M}^d$ embedded in $\mathbb{R}^n$ with the intrinsic dimension $d\ll n$.
For each data point $x_i$, $\mathcal{M}^d$ has a tangent space $T_{x_i}\mathcal{M}$ that includes all vectors at $x_i$ that are tangent to the manifold. 
Let $\{f_1,...,f_n\}$ denote the basis of the data points, which spans source feature  space.

We first depict the manifold structure by weighted $k$NN graph (\Cref{fig:framework}) which approximates the geodesic distance on the manifold~\cite{dong2011efficient}.
Let $(X, E)$ denote the $k$NN graph, where $X=\{x_1,...,x_m\}$ are the data points and $E$ the set of edges $(i,j)$. The edge weight $P_{ij}$ of the $k$NN graph is calculated by the probability distribution on original data ~\cite{mcinnes2018umap} as
\begin{equation}
%\vspace{-1mm}
\label{eq:p}
\begin{split}
\Tilde{P}_{j|i} &= \exp(-(\|x_i - x_j\|-\mathit{dist_i})/\gamma_i)\\
P_{ij} &= \Tilde{P}_{j|i} + \Tilde{P}_{i|j} - \Tilde{P}_{j|i} \Tilde{P}_{i|j}    
\end{split}
%\vspace{-1mm}
\end{equation}
where $\gamma_i$ is adaptive for each $i$ corresponding to the length-scale and $\mathit{dist}_i$ is the distance from $x_i$ to its nearest neighbour. Note that small distance $\|x_i - x_j\|$ between data points $i$ and $j$ induces large similarity $P_{ij}$. 
% More explanation on $P_{ij} xxxxxxxxxx

% For the tangent space $T_{x_i}\mathcal{M}$ of each point $x_i$, we use local PCA to approximate the tangent space ~\cite{singer2012vector,lim2021tangent}. 

Based on the $k$NN graph capturing the topological structure, we retrieve the important features of the data points $X$.
Obviously, the feature information is encoded in the column space of $X$, and the majority of information exists in the principal components~\cite{jolliffe2016principal}. 
Meanwhile, the features are not evenly distributed on the manifold because of curvature: some area on manifold presents plane surface, while some bends more sharply, thus 
we locally calculate principal components of the column space and extract important features.
% Another motivation is that the feature information is encoded in most variable directions. Therefore, we locally compute the directions that maximise the variance to extract important features.

% Most variable directions, refer notebook
% Get PCs as tangent space
% Singular values for ellipsoid

Given a data point $x_i$,  we center its $k$ nearest neighbours as $X_i=[x_{i1}-x_{i},...,x_{ik}-x_{i}]^T \in \mathbb{R}^{k\times n}$. Next, we construct the weight matrix as $W_i = \mathit{diag}(\sqrt{P_{ii_1}},...,\sqrt{P_{ii_k}})\in \mathbb{R}^{k\times k}$. Recall that $P_{ij}$  is the edge weight of $k$NN graph in~\Cref{eq:p}. We assign larger weight to closer neighbours. 

% Assume that we linearly transform the data $X_i$ by $v\in \mathbf{R}^n$. The direction with the largest variance is  
% \begin{equation}
% v_1 = \arg \max_{||v||=1} \{ v^T X_i^T W_i X_i v \} 
% \end{equation}

% Consider the linear transformation of the data $X_i$ by $v\in \mathbf{R}^n$. The directions of transformation with the largest variance are the eigenvectors (corresponding to largest eigenvalues) of the covariance matrix
% \begin{equation}
% \label{eq:covar}
% % v_1 = \arg \max_{||v||=1} \{ v^T X_i^T W_i X_i v \} 
% \mathit{Cov}(X_i) =  X_i^T W_i^T W_i X_i 
% \end{equation}
To locally derive the principal components around data point $x_i$, we apply singular value decomposition (SVD) \footnote{In practice, $k$ is small (\eg $15$), and the extra time cost for SVD is tolerable.} to $\hat{X_i} = W_i X_i$ and get
\begin{equation}
\label{eq:svd}
\hat{X_i} = U_i \Sigma_i V_i^T
% %\vspace{-1mm}
\end{equation}
where the singular values $\Sigma_i = \mathit{diag}(\sigma_{i1},...,\sigma_{ik})$ are in decreasing order and the corresponding right eigenvectors $V_i = [v_{i1},...,v_{ik}]$ span the column space of $\hat{X}_i$ (\Cref{fig:framework}).
The right eigenvectors $V_i$ approximate the tangent space $T_{x_i}\mathcal{M}$  by
the following theorem.
% ~\Cref{theo:local_svd}. 

\begin{theorem} [\cite{singer2012vector,lim2021tangent}]
%\vspace{-3mm}
\label{theo:local_svd}
The right eigenvectors $V_i = [v_{i1},...,v_{ik}]$ by SVD on $\hat{X}_i$ approximately represent an orthonormal basis  for the tangent space $T_{x_i} \mathcal{M}$.
% Tangent space is approximated by local SVD.
%\vspace{-3mm}
\end{theorem}

% The most variable directions are computed by a linear combination of the source features 
% that explains the most variance 
% The span of eigenvectors $\mathit{span}\{V_i\}$ forms the feature (column) space of $\hat{X}_i$. 
The decomposition in~\Cref{eq:svd} illustrates the feature loadings by 
$v^T_{il}=[v^{f_1}_{il},...,v^{f_n}_{il}]$ on the $l$-th principal direction (\Cref{fig:framework}). 
We keep the largest $d$ singular values\footnote{The intrinsic dimension $d$ is locally estimated as the number of singular values that account for most data variability~\cite{singer2012vector,lim2021tangent}.} and the corresponding right eigenvectors, and define the feature importance score for data point $x_i$ as follows:
\begin{definition} [Feature importance]
%\vspace{-2mm}
\label{def:feat_imp}
The feature importance for data point $x_i$ in tangent space $T_{x_i}\mathcal{M}$ is defined as
% in terms of the feature loadings as
\begin{equation}
    \|f^i_h\| = (\sum_{l=1}^d |v^{f_h}_{il}|^2)^{\frac{1}{2}},\  h=1,...,n.
%\vspace{-2mm}
\end{equation}
%\vspace{-3mm}
\end{definition}
The feature importance of source feature $f_h$ is the $l_2$-norm of the corresponding column vector in $V_i^T$, which locally characterises the feature variability around data point $x_i$ on the manifold; features with larger scores are more variable in the ambient space.
% Large feature importance presents more information in this direction. 

\begin{figure}[t]
%\vspace{-2mm}
  \centering
%   \fbox{\rule{0pt}{2in} \rule{0.9\linewidth}{0pt}}
  \includegraphics[width=0.7\linewidth]{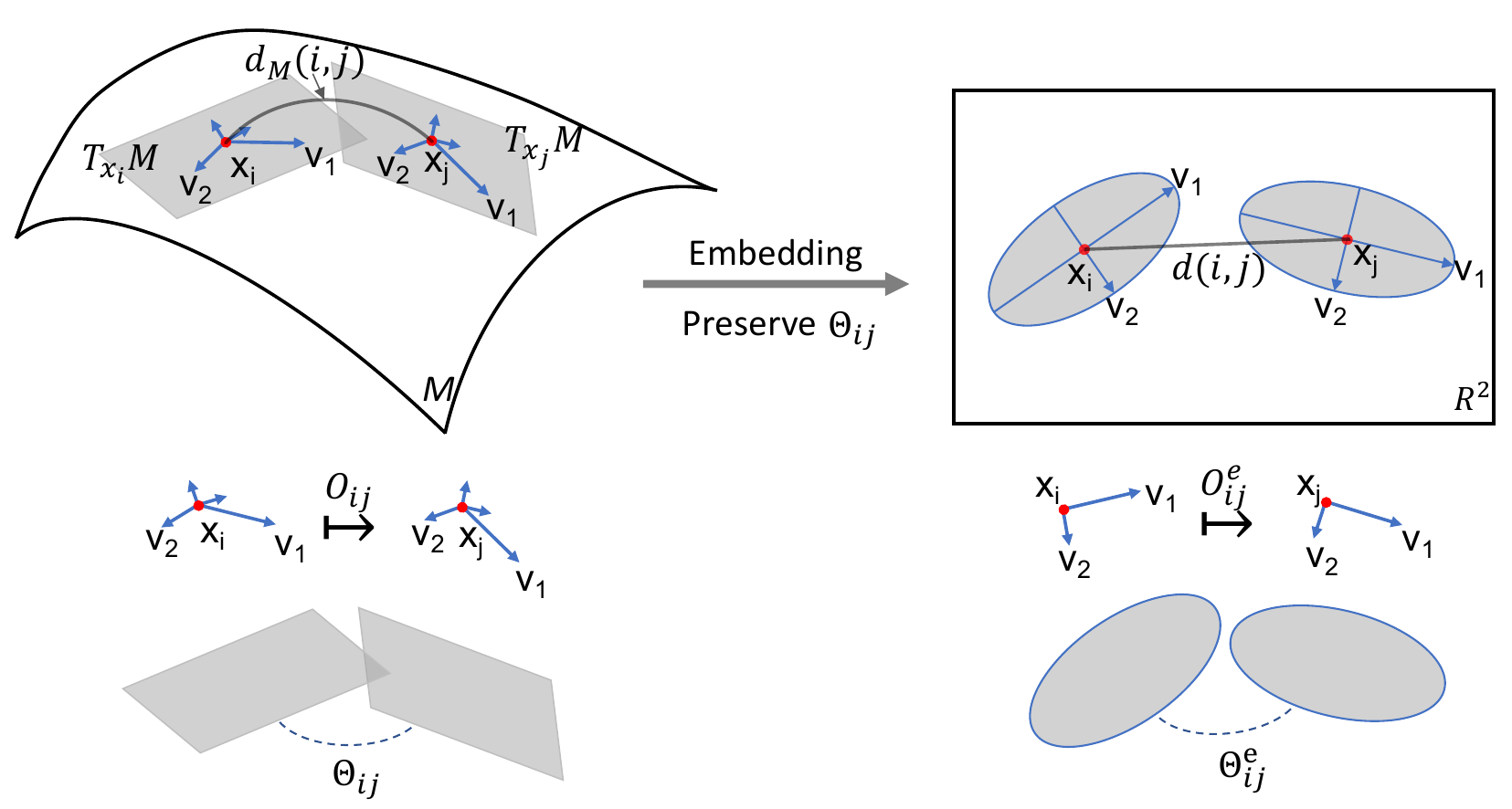}

  \caption{Tangent space embedding.
  The tangent spaces of original data points $x_i$ and $x_j$ are associated with basis vectors ($v_1$ and $v_2$ annotated, on the left);
%   first two principal directions $v_1$ and $v_2$, respectively (left);
  the transformation from $i$ to $j$ consists of translation by geodesic distance $d_{M}(i,j)$ and rotation $O_{ij}$ with the general angle $\Theta_{ij}$  (bottom-left).
  The embedding tangent space (on the right) is computed by preserving the rotation angle $\Theta_{ij}$.
  }
  \label{fig:gauge_emb}
%\vspace{-5mm}
\end{figure}

The definition of feature importance shows that the tangent space locally retains the source feature information. For different data points $x_i$ and $x_j$, the feature importance is separately calculated on the tangent space $T_{x_i}\mathcal{M}$ and $T_{x_j}\mathcal{M}$. Intuitively, tangent space varies at different data points. 
The tangent spaces of two near data points are connected by parallel transport.
We estimate the parallel transport by rotation between the bases of tangent space and refer to it as alignment~\cite{singer2012vector}.
% We characterise the rotation between tangent space by alignment~\cite{singer2012vector}. 
Consider two near data points $x_i$ and $x_j$ with distance $d_\mathcal{M}(i,j)$ on the manifold in~\Cref{fig:gauge_emb}. The orthonormal matrices $V_i, V_j$ approximate the bases of tangent spaces $T_{x_i}\mathcal{M}$ and $T_{x_j}\mathcal{M}$, respectively. The transformation from $x_i$ to $x_j$ involves translation and rotation operation. We calculate the translation by distance $d_\mathcal{M}(i,j)$, and the rotation by optimal alignment $O_{ij}$ which is
\begin{equation}
\label{eq:align}
    O_{ij} = \arg \min_{O\in O(d)}\|O - V_j V_i^T\|_F.
\end{equation}
Numerically, the alignment $O_{ij}$ is computed by the SVD of $V_j V_i^T=U'\Sigma' V'^T$ and $O_{ij}=U'V'^T$.
% Therefore, the transformation from tangent space $T_{x_i}\mathcal{M}$ to $T_{x_j}\mathcal{M}$ is
We define the transformation as:
\begin{equation}
\label{eq:transf}
\begin{split}
\Gamma_{i\xrightarrow{} j}: T_x\mathcal{M} &\xrightarrow{} T_x\mathcal{M}\\
X &\mapsto O_{ij} X + d_{\mathcal{M}}(i,j).
\end{split}
%\vspace{-2mm}
\end{equation}
% \begin{definition} [Transformation]
% \label{def:trans}
% The transformation from tangent space $T_{x_i}\mathcal{M}$ to $T_{x_j}\mathcal{M}$ is
% \begin{equation}
% \label{eq:transf}
% \begin{split}
% \Gamma_{i\xrightarrow{} j}: T_x\mathcal{M} &\xrightarrow{} T_x\mathcal{M}\\
% X &\mapsto O_{ij} X + d_{\mathcal{M}}(i,j)
% \end{split}
% \end{equation}
% \end{definition}
The transformation $\Gamma_{i\xrightarrow{} j}$ represents the connection of data points $i$ and $j$ on both topological space (by distance $d_{\mathcal{M}}(i,j)$) and  tangent space (by alignment $O_{ij}$). The conventional nonlinear dimensionality reduction methods focus on the former connection by preserving pairwise distance~\cite{coifman2005geometric,van2008visualizing,mcinnes2018umap}, which succeed in maintaining the topological structure, whereas the tangent space is missing in these methods, resulting in no source features preserved in low-dimensional embedding space.

We aim to preserve the topological structure as well as tangent space to reveal source features in low-dimensional space. 
We first embed the tangent space by preserving the alignment (\Cref{eq:align}), followed by maintaining the local distance along the embedding tangent space in low-dimensional space.

% To preserve the source feature information in low-dimensional space, we need to first embed the tangent space. 

% The source feature information is locally encoded in the column space of $\hat{X}_i$.

% The right eigenvectors $V_i$ also approximate the tangent space $T_{x_i}\mathcal{M}$ at point $x_i$.

% We aim to preserve the source features in low-dimensional space. 

% Assume the manifold $\mathcal{M}^d$ is sampled at data points $\{x_1,...,x_m\}\subset \mathbb{R}^n $
% Original space:
% knn-graph -- Local PCA for tangent space

% Optical alignment -- parallel transport

\subsection{Tangent space embedding}
\label{subsec:gauge_emb}
The feature space of data points $X=\{x_1,...,x_m\}$ is locally depicted by the tangent space $T_x{\mathcal{M}}$ (\Cref{def:feat_imp}). We project the tangent space to low-dimensional space by preserving the alignment between tangent spaces (\Cref{fig:gauge_emb}).

For each point $x_i$, the tangent space $T_{x_i}\mathcal{M}$ is estimated by the $d$ largest right eigenvectors $V_i=[v_{i1},...,v_{id}]$ in \Cref{eq:svd}.
The overall tangent space $T_x\mathcal{M}$ is approximated as $V=[V_1;...;V_m]\in \mathbb{R}^{m\times n\times d}$, which is a $3$-order tensor demonstrating $m$ data points with $n\times d$-dimensional feature space. 

% We extend the scheme of $k$NN graph embedding \cite{coifman2005geometric,van2008visualizing,mcinnes2018umap} to tensor embedding for tangent space $V\in \mathbb{R}^{m\times n\times d}$.

We consider the tensor embedding on tangent bundle $\mathcal{M} \times T_x \mathcal{M}$,
where the topological structure of $\mathcal{M}$ is calculated by $k$NN graph and tangent space $ T_x \mathcal{M}$ is by local SVD.
For two data points $x_i$ and $x_j$ with corresponding tangent space basis $V_i$ and $V_j$ respectively, we define the consistence degree~\cite{deng2020low} between the tangent spaces $T_{x_i}\mathcal{M}$ and $T_{x_j}\mathcal{M}$ as
\begin{equation}
\label{eq:cos}
\cos{\Theta_{ij}} = \frac{\langle V_i, V_j\rangle_F}{\|V_i\|_F \|V_j\|_F}
\end{equation}
where $\langle V_i, V_j\rangle_F = \mathit{tr}(V_i V_j^T)$  denotes the Frobenius inner product and $\|\cdot\|_F$ is Frobenius norm. \Cref{eq:cos} presents the cosine similarity  between tangent space $T_{x_i}\mathcal{M}$ and $T_{x_j}\mathcal{M}$, which induces a general angle $\Theta_{ij}$ in~\Cref{fig:gauge_emb}.
Note that the above consistence degree quantitatively characterises the alignment in ~\Cref{eq:align}.
% For an edge $(i,j)$,  we compute the matrix similarity $\mathbf{P}_{ij}\in \mathbb{R}^{d\times d}$ of the two tangent space $T_{x_i}\mathcal{M}$ and $T_{x_j}\mathcal{M}$ as the alignment $O_{ij}\in \mathbb{R}^{d\times d}$ in \Cref{eq:align}.
% Note that the entry $O_{ij}(s,t)$ of alignment represents the transformation from $t$-th direction $v_{it}$ of tangent space $T_{x_i}\mathcal{M}$ to the $s$-th direction $v_{js}$ of $T_{x_j}\mathcal{M}$. 
We define the probability distribution of the consistence degree on original tangent space as
\begin{equation}
\label{eq:sim_o_tangent}
\begin{split}
% \mathbf{P}_{j|i}(s,t) &= \exp(-|1-O_{ij}(s,t)|/\tau_i),\\
%     s,t &= 1,...,d
\mathbf{P}_{j|i}(\Theta) = \exp(-(|1-\cos{\Theta_{ij}}|-\mathit{dist'_i})/\gamma'_i)
\end{split}
\end{equation}
% where $\tau_i$ is chosen adaptively and 
where $1-\cos{\Theta_{ij}}$ is the cosine distance and we set normalization to $\mathbf{P}_{j|i}(\Theta)$ similar as \Cref{eq:p} to get $\mathbf{P}_{ij}(\Theta)$. 
We focus on preserving the pairwise cosine distance of tangent spaces in low-dimensional embedding space.
% Note that the alignment score $O_{ij}(s,t)$ corresponds to cosine similarity between two tangent space.

Consider projecting the manifold to $d'$-dimensional space ($d'\leq d\ll n$). For each data point $x_i$, let $V_i^e = [v^e_{i1},...,v^e_{id'}] \in \mathbb{R}^{d'\times d'}$ denote the embedding tangent space. 
Similarly, we define the cosine similarity between $V_i^e$ and $V_j^e$ as
\begin{equation}
\label{eq:cos_e}
\cos{\Theta_{ij}^e} = \frac{\langle V_i^e, V_j^e\rangle_F}{\|V_i^e\|_F \|V_j^e\|_F}
\end{equation}
with the angle $\Theta_{ij}^e$ in~\Cref{fig:gauge_emb}.
% The alignment from $x_i$ to $x_j$ in embedding space is computed as 
% \begin{equation}
% \label{eq:align_e}
% \begin{split}
%         O^e_{ij} &= (V_j^e)^T V_i^e,\\
%         &(V_i^e)^T V_i^e = (V_j^e)^T V_j^e = I \in \mathbb{R}^{d'\times d'}. 
% \end{split}
% \end{equation}
We aim to learn the tangent space embedding $V^e = \{V_1^e,...,V_m^e\}$ to make $V^e$ have similar distribution on cosine distance to original space by~\Cref{eq:sim_o_tangent}.
% Formally, we compute the probability distribution on original tangent space as
% \begin{equation}
% \label{eq:sim_tang_o}
% \mathbf{P}_{j|i}(\Theta) = \exp(-\|1-\cos{\Theta_{ij}}\|/\tau_i),\\
% \end{equation}
For the probability distribution on the embedding tangent space, we inherit the heavy-tailed distribution from~\cite{van2008visualizing,mcinnes2018umap} as: 
% The probability distribution for pair $(i, j)$ on embedding space is
\begin{equation}
\label{eq:sim_e_tangent}
\begin{split}
% \mathbf{Q}_{ij}(s,t) &= (1 + ad_{ij}^{2b}(s,t))^{-1}, \\
% s,t &= 1,...,d'
\mathbf{Q}_{ij}(\Theta^e) &= (1 + ad_{ij}^{2b}(s,t))^{-1}
\end{split}
\end{equation}
where $ d_{ij}=1-\cos{\Theta_{ij}^e}$ represents the cosine distance between the embedding tangent spaces $V_i^e$ and $V_i^e$ and $a, b$ are shape parameters. 

To preserve the alignment in embedding tangent space, we minimize the difference between $\mathbf{P}$ and $\mathbf{Q}$ by Kullback–Leibler divergence:
% For the $s$-th and $t$-th directions in tangent space, the difference by Kullback–Leibler divergence (or cross-entropy loss) is:
\begin{equation}
\begin{split}
    % \mathit{KL}(\mathbf{P}(s,t)|| \mathbf{Q}(s,t)) &= \\
    % - \sum_{ij} \mathbf{P}_{ij}(s,t)&(\log \mathbf{P}_{ij}(s,t) - \log \mathbf{Q}_{ij}(s,t)),
    \mathit{KL}(\mathbf{P}|| \mathbf{Q}) =
    - \sum_{ij} \mathbf{P}_{ij}(\log \mathbf{P}_{ij} - \log \mathbf{Q}_{ij}),
\end{split}
\end{equation}
% then, the difference between probability distributions $\mathbf{P}$ and $\mathbf{Q}$:
% \begin{equation}
% \label{eq:kl}
% \mathit{KL}(\mathbf{P}|| \mathbf{Q}) = \sum_{s,t} \mathit{KL}(\mathbf{P}(s,t)|| \mathbf{Q}(s,t)),
% \end{equation}
and we optimize the embedding feature space $V^e=[V_1^e;...;V_m^e]$ to minimize the loss function by stochastic gradient descent (SGD). 

% \noindent \textbf{Remark.} In practice, for two-dimensional data visualization, we learn embedding tangent space of point $i$ as $V_i^e=[v_{i1}^e, v_{i2}^e]$. Since $(v_{i1}^e)^T v_{i2}^e =0$, it is sufficient to learn the embedding $v_{i1}^e$ of the first direction, thus the tensor embedding problem degrades to matrix embedding which preserves the pairwise angular of the first directions induced by alignment in~\Cref{eq:align,eq:align_e}.

% $V_i^e = \big(\begin{smallmatrix}
%   \cos{\theta_i} & \sin{\theta_i}\\
%   -\sin{\theta_i} & \cos{\theta_i}
% \end{smallmatrix}\big)$,
% which is a collection of rotation matrices. Since there is one parameter $\theta_i$, 

% In practice, 2-dimensional; rotation matrix

% Frame for features.
The embedding tangent space $V^e=[V_1^e;...;V_m^e]$ equips the topological space with a frame for source features (on the right of~\Cref{fig:framework}). The feature loadings (dashed arrows) are derived from the top $d'$ eigenvectors of original tangent space $V$, and the feature importance is illustrated by the arrows' length.

The embedding tangent space $V^e$ also presents the directions along which the data points are locally distributed in low-dimensional space. Note that the directions of tangent space $T_{x_i}\mathcal{M}$ are weighted by the singular values $[\sigma_{i1},...,\sigma_{id}]$ (\Cref{eq:svd}), which forms a hyperellipsoid showing anisotropic density (top-middle of~\Cref{fig:framework}). Thus, we will project the original data points along the local weighted directions to preserve the anisotropic density in low-dimensional space.
% Based on the embedding tangent space, we project the data points of original space along the weighted directions to preserve 

% Optimize the embedding coordinates by SGD.

% Get the directions of tangent space in embedding space

% Each direction has weight $\sigma_i$

% Project data point by direction and weight

% high-dim tangent space
% low-dim tangent space

% Entries in $O_{ij}$ denote the inner product (cosine sim) of original local axies.
% Distance based dimensionality reduction to get gauge embedding

% low-dim space:
% Embed the gauge to preserve parallel transport

\subsection{Anisotropic projection}
In this section, we compute the $d'$-dimensional ($d'\leq d\ll n$) embedding points $Y=\{y_1,...,y_m\}$ under the frame of embedding tangent space $V^e$.
% shown on the right of~\Cref{fig:gauge_emb}.
We have constructed the tangent space $V^e=[V_1^e;...;V_m^e]\in \mathbb{R}^{m\times d'\times d'}$ in low-dimensional embedding space (\Cref{fig:gauge_emb}), and each local space $V_i^e$ is associated with weights by the singular values $\Sigma_i^e = \mathit{diag}(\sigma_{i1},...,\sigma_{id'})$. This tangent space locally portrays an anisotropic projection of data points $X = \{x_1,...,x_m\}$ to low-dimensional space.

We first calculate the  $1$-skeleton topological structure in low-dimensional space by the heavy-tailed distribution~\cite{mcinnes2018umap}:
\begin{equation}
\label{eq:q}
Q_{ij} = (1 + a\|y_i - y_j\|^{2b})^{-1}
\end{equation}
where $||y_i - y_j||$ denotes the (Euclidean) distance in embedding space, and $a,b$ are shape parameters.
% Note that the distance shows the interaction on embedding space in~\Cref{def:trans}.
We preserve the distance by minimizing the difference between distribution $Q_{ij}$ and $P_{ij}$ (\Cref{eq:p}) by cross-entropy loss
\begin{equation}
\label{eq:ce}
\mathit{CE}(P||Q) = -\sum_{ij} P_{ij}\log Q_{ij} + (1-P_{ij})\log (1-Q_{ij})
\end{equation}

Consider the low-dimensional data points $y_i$ with the embedding frame $V_i^e = [v^e_{i1},...,v^e_{id'}]\in \mathbf{R}^{d'\times d'}$. We center its $k$ nearest neighbours as $Y_i = [y_{i1}-y_i,...,y_{ik}-y_i]^T \in \mathbb{R}^{k\times d'}$, and project $Y_i$ to its tangent space $V_i^e$ as $\hat{Y}_i = Y_i V_i^e$. 
% Let $y_j$ be one of $y_i$'s neighbours. 
For one of $y_i$'s neighbours, say $y_j$,
the distance $||y_i - y_j||$ is invariant under the rotation $V_i^e$, which makes $\hat{Y}_i$ fit~\Cref{eq:q}.
% \begin{equation}
% \begin{split}
%     ||(y_j - y_i)^T V_i^e||^2 &= (y_j - y_i)^T V_i^e (V_i^e)^T (y_j - y_i) \\
%     % &= (y_j - y_i)^T (y_j - y_i)
%     &=||y_i - y_j||^2,
% \end{split}
% \end{equation}
% since $V_i^e (V_i^e)^T = I$ by orthogonality, 

Recall that the original tangent space $V_i$ with weight $\Sigma_i = \mathit{diag}(\sigma_{i1},...,\sigma_{id})$ forms a hyperellipsoid (top-middle of~\Cref{fig:framework}) with radius $R_{il}^o = \sigma_{il}^2, l=1,...,d$\footnote{We use squared distance (variance) because of better empirical performance.}.
We reshape the embedding data points $\hat{Y}_i = Y_i V_i^e$ such that the volume of the point cloud $\hat{Y}_i$ agrees with the original hyperellipsoid.
The radius of $\hat{Y}_i$ in direction $v_{il}^e$ ($l = 1,...,d'$) is 
\begin{equation}
\label{eq:r_e}
\begin{split}
    R_{il}^e = \frac{1}{\sum_j Q_{ij}} \sum_j Q_{ij} \|(y_j - y_i) v_{il}^e\|^2, 
\end{split}
\end{equation}
where $R_{il}^e$ ($l=1,...,d'$) results from the anisotropic projection.
% we assign more weights to closer points. 
The local hyperellipsoid volume of original and embedding space is 
$c(d)\prod_{l=1}^{d} R_{il}^o = c(d')\prod_{l=1}^{d'} R_{il}^e$, where $c$ denotes a constant related to dimensions. 
We reframe this equation by logarithm and get
\begin{equation}
    r_{i1}^o +...+r_{id'}^o + c = r_{i1}^e +...+r_{id'}^e
\end{equation}
where $r_{il}^o = \log R_{il}^o$ and $r_{il}^e = \log R_{il}^e$.
To preserve the volume in embedding space, it is sufficient to set $r_{il}^e = \beta r_{il}^o + \alpha,i=1,...,m$ in the $l$-th ($l=1,...,d'$) direction.
We measure the goodness of fitting this relationship by correlation coefficient
\begin{equation}
\mathit{Corr}(r_{l}^o, r_{l}^e) =  \frac{\mathit{Cov}(r_{l}^o, r_{l}^e)}{(\mathit{Var}(r_{l}^o) \mathit{Var}(r_{l}^e))^{1/2} }
\end{equation}
and for all directions, we have
\begin{equation}
\label{eq:corr_1}
\mathit{Corr}(r^o, r^e) = \sum_{l=1}^{d'} \mathit{Corr}(r_{l}^o, r_{l}^e)
\end{equation}
We combine the loss function for preserving $1$-skeleton structure (\Cref{eq:ce}) and correlation coefficient for anisotropic density (\Cref{eq:corr}) to get the overall loss function
\begin{equation}
\label{eq:loss}
\mathcal{L} = \mathit{CE}(P||Q) - \lambda \mathit{Corr}(r^o, r^e)
\end{equation}
where $\lambda$ determines the relative importance of anisotropic density preservation.
We use SGD to minimize the above loss function to get $d'$-dimensional embedding coordinates $Y = \{y_1,...,y_m\}$.

Therefore, the embedding coordinates $Y = \{y_1,...,y_m\}$ by the anisotropic projection not only preserve the topological structure of local similarity and density, but also encapsulate the embedding tangent space $V^e$ which locally demonstrates the source features (on the right of~\Cref{fig:framework}). 
% In addition, the anisotropic projection enable the method to maintain the density of original data.

%\vspace{1mm}
\noindent \textbf{Training.}
% We minimize the loss function to get the embedding coordinates $Y=\{y_1,...,y_m\}$ by SGD.
The calculation details for SGD are in Supplementary. 
In practice, we set the (hyper)parameters same as UMAP~\cite{mcinnes2018umap} including number of neighbours, number of iterations and the 'min-dist' parameter. There are two additional parameters: the weight $\lambda \leq 0$ for the anisotropic density preservation and the fraction $q\in [0,1]$ of iterations that considers tangent space embedding. We use $15$ neighbours, $500$ epochs, $q=0.3$ and $\lambda = 0.5$. 
We include tuning of $\lambda$ in Supplementary (\cref{fig:lambda}).

\section{Experiments}
%\vspace{-1mm}
\label{sec:experiments}
We evaluate featMAP in interpreting MNIST digit classification as well as Fashion MNIST and COIL-20 object detection by using source features in embedding space. 
% diverse datasets that cover domains such as digits, objects and biological data.
We also apply featMAP to interpreting MNIST adversarial examples, showing that our method featMAP uses feature importance to explicitly explain the misclassification after adversarial attack.
In addition, we show that featMAP maintains the original density by anisotropic projection.
In the end, we compare featMAP with state-of-the-art algorithms by local and global structure preservation metrics.

%\vspace{-1mm}
\subsection{Datasets and evaluation metrics}
%\vspace{-2mm}
We perform the evaluation on various datasets including
standard MNIST~\cite{lecun1998mnist}, Fashion MNIST~\cite{xiao2017fashion}, COIL-20~\cite{nene1996columbia}, Cifar10~\cite{krizhevsky2009learning}, single cell RNA-seq~\cite{liu2021time}, and MNIST adversarial examples~\cite{goodfellow2014explaining}.
We compare with representative methods for dimensionality reduction and visualization, including t-SNE~\cite{van2008visualizing}, FIt-SNE~\cite{van2014accelerating}, h-NNE~\cite{sarfraz2022hierarchical}, UMAP~\cite{mcinnes2018umap}, triMAP~\cite{amid2019trimap}, PaCMAP~\cite{wang2021understanding}, densMAP~\cite{narayan2021assessing} and spaceMAP~\cite{zu2022spacemap} in terms of both local and global structure preservation~\cite{espadoto2019toward, wang2021understanding}. 
Local structure preservation metrics include $k$NN accuracy, trustworthiness~\cite{venna2006local} and continuity~\cite{venna2006visualizing}. 
% $k$NN accuracy computes the classification accuracy of $k$NN classifier on low-dimensional embedding data; we use $10$-fold cross-validation to measure the $k$NN accuracy by varying $k$ from $2$ to $64$.
% Trustworthiness expresses to what extent the local topological structure is retained by computing true neighbour rate, while continuity measure the local structure of the embedding by missing neighbour rate.  
Global structure preservation metrics cover Shepard goodness, normalized stress~\cite{joia2011local} and triplet centroid accuracy~\cite{wang2021understanding}.
% Shepard goodness is calculated by the Spearman rank correlation of Shepard diagram which is the scatter plot of pairwise distance of all data points in embedding space versus original space. 
% Normalized stress measures the preservation of distance from original space to embedding space.
% Triplet centroid accuracy measures the percentage of triplets whose relative distance maintains the relative order in original and embedding space.
The details to calculate these metrics are found in ~\cite{espadoto2019toward, wang2021understanding}.

%\vspace{-1mm}
\subsection{Results}
%\vspace{-1mm}

% We first show that featMAP preserves source features, thus enabling the interpretability of the nonlinear dimensionality reduction results. We then apply featMAP to the adversarial examples of MNIST, using feature importance to explain how the images are attacked from original classification to adversarial classification.

\subsubsection{FeatMAP preserving source features}
\begin{figure}[ht]
%\vspace{-2mm}
  \centering
%   \fbox{\rule{0pt}{2in} \rule{0.9\linewidth}{0pt}}
  \includegraphics[width=0.7\linewidth]{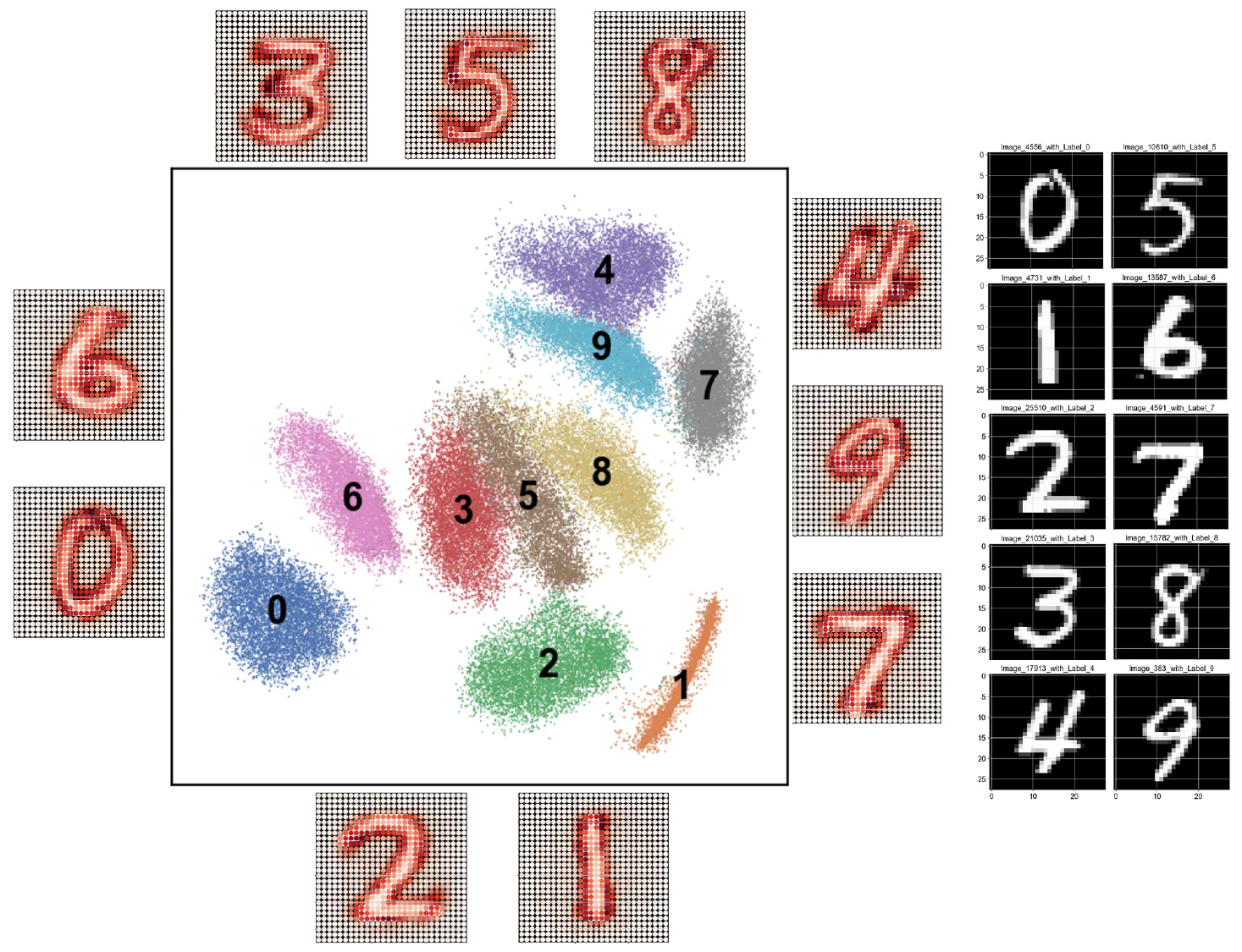}

  \caption{FeatMAP on MNIST showing feature importance. 
  FeatMAP embeds MNIST to $2$-dimensional space with $10$ different clusters (left).
  Digit images are randomly selected from each cluster illustrating the feature importance with corresponding original images (right).
  Darker red means larger feature importance. 
  }
  \label{fig:feat_imp}
%\vspace{-5mm}
\end{figure}

\begin{figure}[ht]
%\vspace{-2mm}
  \centering
%   \fbox{\rule{0pt}{2in} \rule{0.9\linewidth}{0pt}}
  \includegraphics[width=0.6\linewidth]{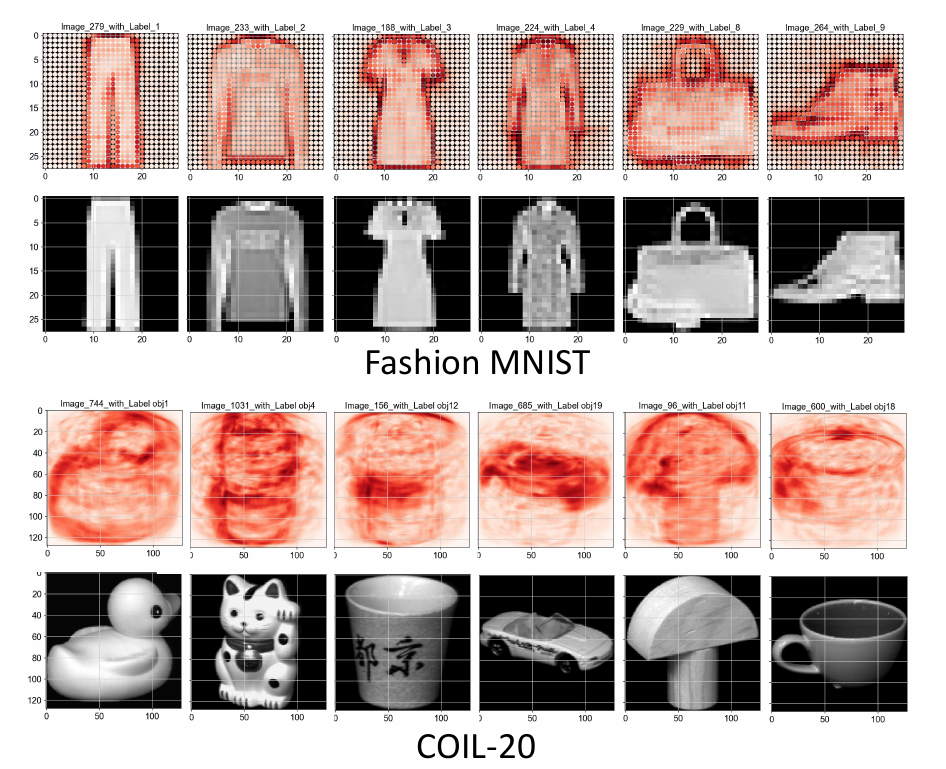}
  \caption{FeatMAP on Fashion MNIST and COIL-20 showing feature importance. For each dataset, the upper part is the saliency map by feature importance and the bottom are the original images.}
  \label{fig:feat_imp_coil}
  %\vspace{-6mm}
\end{figure}

%\vspace{-2mm}
% \noindent \textbf{FeatMAP preserving source features.} 
We demonstrate that featMAP enables interpretable dimensionality reduction by preserving source features on MNIST digit classification as well as Fashion MNIST and COIL-20 object detection.
\Cref{fig:feat_imp} clearly shows the clusters of different digit groups using featMAP. 
We randomly choose one data point from each cluster and show the feature importance as saliency map for the corresponding images. 
We find that the pattern by feature importance explicitly detects the edge of each digit.
Similarly, we apply featMAP to datasets Fashion MNIST and COIL-20 in~\Cref{fig:feat_imp_coil}. The important features succeed in detecting the objects of different categories in both datasets. We also find that for COIL-20 dataset, the rotation patterns (the lighter red curves) are revealed in the saliency map, which originates from generating the dataset COIL-20 by rotating those objects with a fixed camera~\cite{nene1996columbia}.

We further illustrate the interpretability of featMAP by showing the feature loadings and feature importance in~\Cref{fig:feat_load}. 
Specifically, we annotate the top-$10$ most important pixel features. 
These pixels mainly appear in the corner angles of the digit object, indicating that these features locally dominate the image.
This phenomenon is also observed in the saliency map of digits in~\Cref{fig:feat_imp}.

% object detection
% adversarial examples

% \begin{figure*}[b]
%   \centering
% %   \fbox{\rule{0pt}{2in} \rule{0.9\linewidth}{0pt}}
%   \includegraphics[width=\linewidth]{Figures/fig_metric.pdf}

%   \caption{FeatMAP on MNIST (digit $1$) showing density preservation.
%   xxxxxxx
%   xxxxxxx
%   }
%   \label{fig:metric}
% \end{figure*} 

\begin{figure*}[h]
%\vspace{-2mm}
  \centering
%   \fbox{\rule{0pt}{2in} \rule{0.9\linewidth}{0pt}}
  \includegraphics[width=0.8\linewidth]{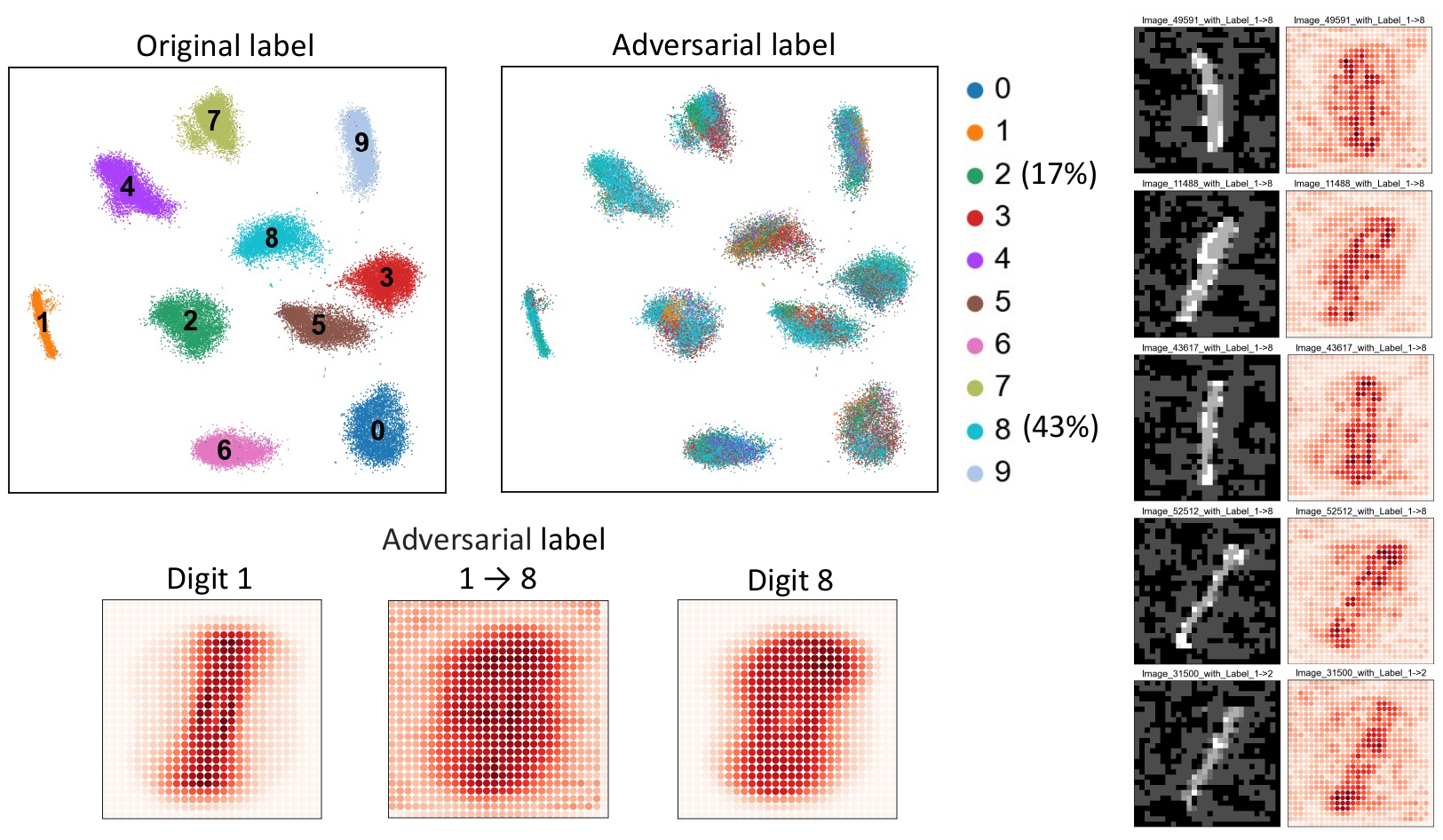}
  \caption{ FeatMAP on MNIST adversarial examples. 
 FeatMAP embeds MNIST adversarial examples to $2$-dimensional space with original and adversarial labels (top-left).
 Five randomly selected adversarial examples from original digit $1$ group are shown on the right.
 The bottom-left shows the average feature importance of adversarial examples misclassified from $1$ to $8$ and the corresponding digits $1$ and $8$ in original data before attack.
}
  \label{fig:adver}
%\vspace{-6mm}
\end{figure*}   
%\vspace{-4mm}
\subsubsection{FeatMAP interpreting adversarial examples}
%\vspace{-2mm}
% \noindent \textbf{FeatMAP interpreting adversarial examples.}
We show that featMAP interprets MNIST adversarial examples by explaining the misclassification after adversarial attack.
% We generate adversarial examples by Fast Gradient Sign Attack (FGSM)~\cite{goodfellow2014explaining} on MNIST with $\epsilon=0.3$ and prediction accuracy $0.08$.
We use Fast Gradient Sign Attack (FGSM)~\cite{goodfellow2014explaining} to synthesize fake images of MNIST with $\epsilon=0.3$ and prediction accuracy $0.08$, to fool the  classifier LeNet~\cite{lecun1998gradient}. 
Formally, the adversarial example $x'\in \mathbb{R}^n$ satisfies $\|x' - x\| \leq \epsilon$, where $x$ is the original image, and the predicted label $f(x')\neq f(x)$.
The results are included in~\Cref{fig:adver} and Supplementary~\Cref{fig:adv_other}.

We plot the adversarial examples by featMAP in~\Cref{fig:adver}. The top-left part shows the original and adversarial labels.
We find that the clustering structures of the adversarial examples are quite similar to original clusters in~\Cref{fig:feat_imp}, because the adversarial examples fall within $x$'s $\epsilon$-neighbour ball ($\epsilon = 0.3$) which locally resembles the topological structure of original examples.
FGSM successfully fools the classifier with $43\%$ data misclassified as digit $8$.
We randomly select five images from the cluster with original label digit $1$, and illustrate the adversarial images  with feature importance on the right of~\Cref{fig:adver}. 
The adversarial images are identified as digit $1$ by the naked eye, while the saliency map reveals other important features besides the edge features of digit $1$. 
% detected in~\Cref{fig:feat_imp}.
We average the feature importance of the adversarial examples misclassified from digit $1$ to $8$, and compare it with the original average feature importance of digit $1$ and $8$ (bottom-left in~\Cref{fig:adver}). We find that the important feature patterns in adversarial examples are more similar to the pattern in original label $8$ than $1$, which evidently explains the misclassification after attack. 

% the Fast Gradient Sign Attack (FGSM), to fool an MNIST classifier
% to synthesize a fake image perceptually similar to original image but that
% can mislead the classifier to give wrong prediction results

% Adversarial attack constraints $||x - x'|| \leq \epsilon$, thus cluster structure same;

% But labels are different

% The reason for different labels 

%\vspace{-4mm}
\subsubsection{FeatMAP preserving density}
\begin{figure}[h]
%\vspace{-2mm}
  \centering
%   \fbox{\rule{0pt}{2in} \rule{0.9\linewidth}{0pt}}
  \includegraphics[width=0.5\linewidth]{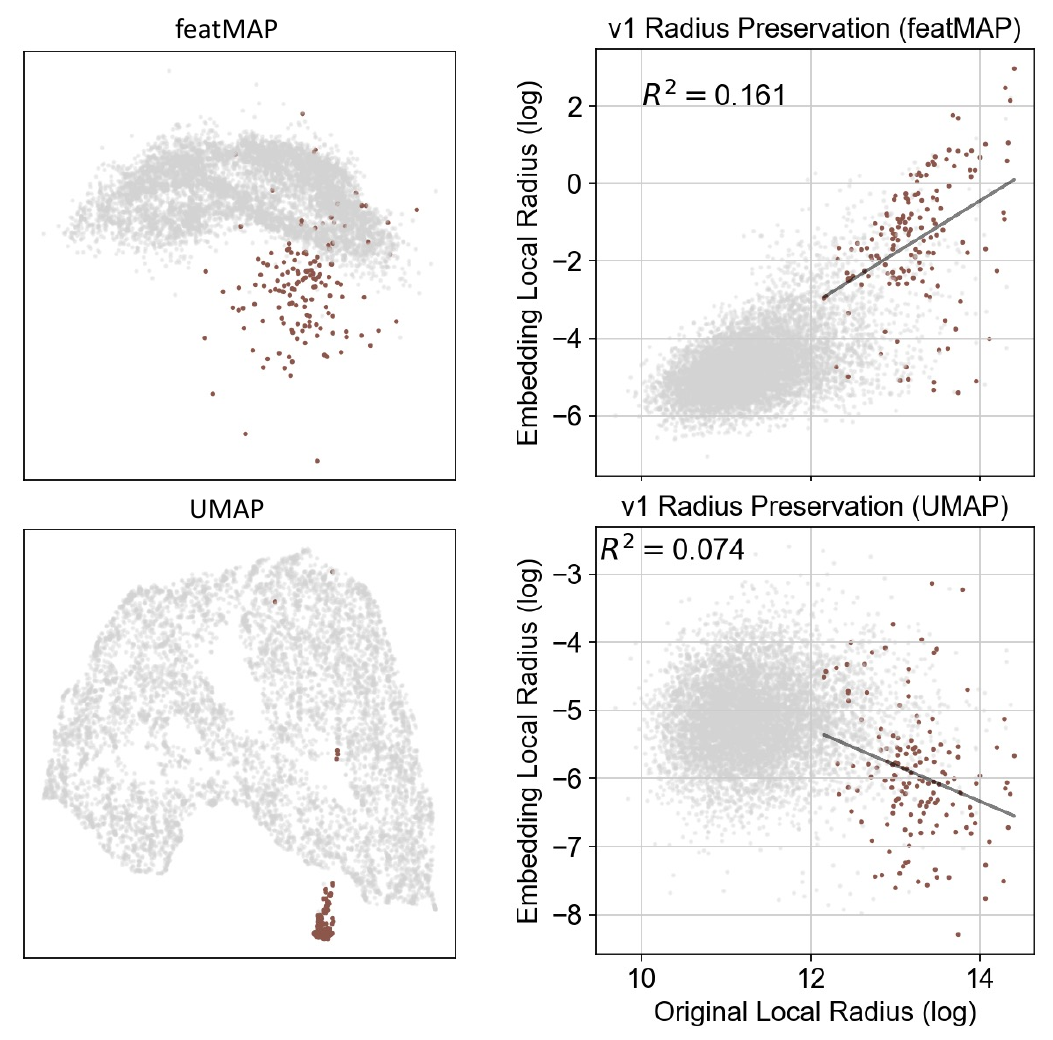}

  \caption{FeatMAP on MNIST (digit $1$) showing density preservation.
  For one subgroup (colour in brown), featMAP (top) preserves the local radius against UMAP (bottom) by positive correlation between embedding and original local radius;
  featMAP correctly illustrates the sparse pattern compared to UMAP.}
  \label{fig:density}
  %\vspace{-6mm}
\end{figure}

%\vspace{-2mm}
% \noindent \textbf{FeatMAP preserving density.}
We illustrate that featMAP maintains original data density in two-dimensional plot. 
% in~\Cref{fig:density} and Supplementary. 
% which is also reported by densMAP~\cite{narayan2021assessing}. 
We apply featMAP to the digit $1$ group of MNIST dataset and plot the correlation of local radius between embedding and original space.
FeatMAP presents larger local radius correlation than UMAP (in Supplementary~\Cref{fig:den_1}),
indicating that featMAP performs better in local density preservation.
% We cluster the dataset by~\cite{traag2019louvain}.
% in Figure xx (Supplementary). 
Particularly, we cluster the dataset and find that the subgroup (in brown colour) is significantly more sparse in featMAP than UMAP in~\Cref{fig:density}.
We claim that featMAP correctly reveals this sparse pattern because featMAP shows positive correlation of local radius between embedding and original space.
% We cluster the dataset and find that featMAP correctly shows the positive correlation of local radius between embedding and original space for the subgroup (in brown colour) in ~\Cref{fig:density}.
% The plot of this subgroup by featMAP presents more sparse pattern than UMAP.
% the subgroup (in brown colour) is significantly more sparse in featMAP than UMAP in~\Cref{fig:density}.
% and local radius correlation in featMAP is larger than UMAP in~\Cref{fig:density}. 
We further demonstrate the digit images from this subgroup in Supplementary~\Cref{fig:sparse}, which exhibit diverse handwritten patterns and agree with the sparse pattern in featMAP.

% Note that the cluster of digit $1$ is more compact than the others. 

%\vspace{-1mm}
% Remark that the cluster for digit $1$ exhibits denser structure than UMAP  
\subsection{Quantitative comparison with state-of-the-art}
%\vspace{-1mm}
We compare featMAP with representative dimensionality reduction methods. The two-dimensional visualization plots are in Supplementary~\Cref{fig:overall_plot}.
We report the quantitative comparison results in~\Cref{tbl:metric}.
FeatMAP illustrates comparable results on both local and global metrics. 

\begin{table*}[htbp]
%\vspace{-2mm}
 \caption{Quantitative comparison. $M_t$, $M_c$ and $M_k$ indicate the local metrics as trustworthiness, continuity and $k$NN accuracy, respectively; $M_s$, $M_\sigma$ and $M_{ct}$ denote the global metrics as Shepard goodness, normalized stress and centroid triplet accuracy, respectively.}
 \label{tbl:metric}
    \begin{subtable}{0.49\textwidth}
         \includegraphics[width=\linewidth]{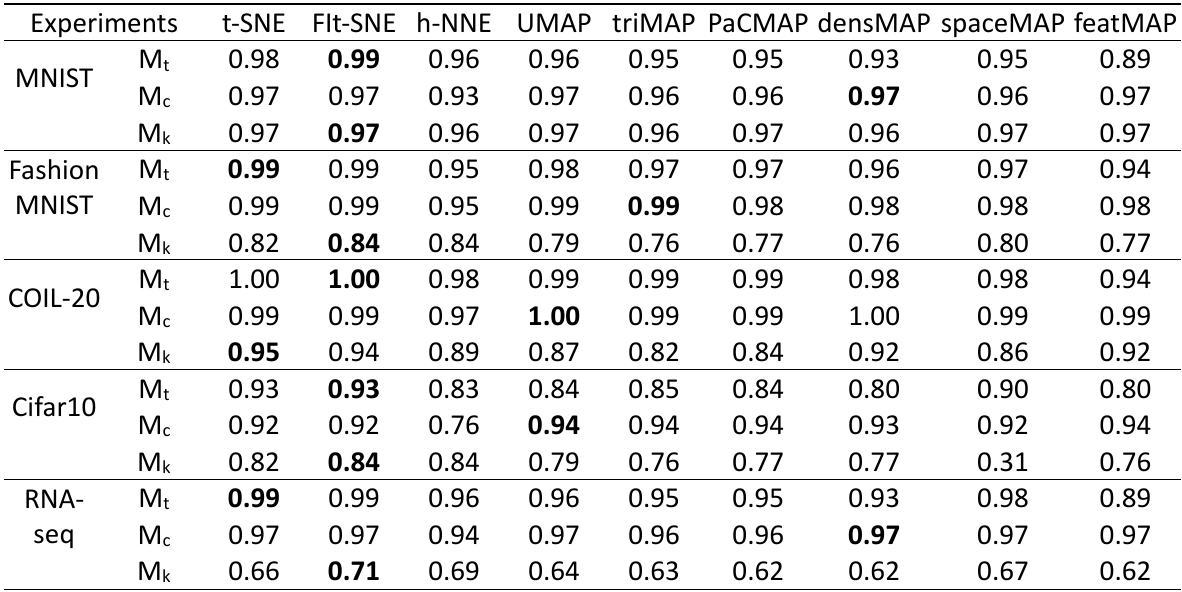}
         \caption{Local metrics}
    \end{subtable}
    \hfill
    \begin{subtable}{0.49
    \textwidth}
         \includegraphics[width=\linewidth]{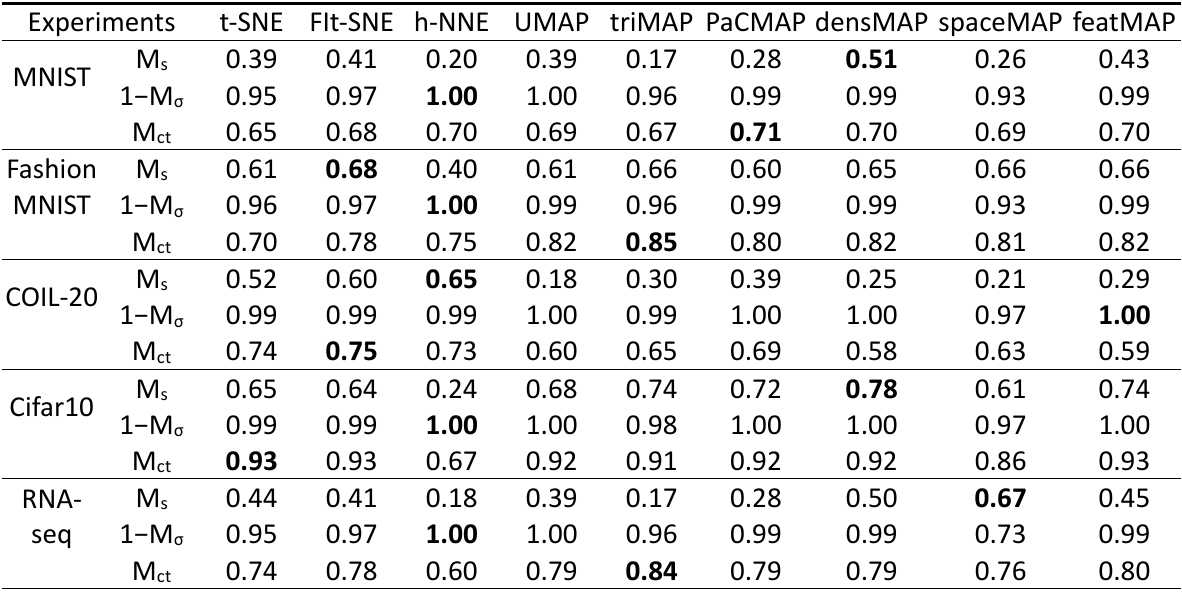}
         \caption{Global metrics}
    \end{subtable}
%\vspace{-6mm}
\end{table*}

% \section{Discussion}
% \label{sec:discussion}

%\vspace{-2mm}
\section{Conclusion}
\label{sec:conclusion}
%\vspace{-2mm}

We propose featMAP enabling the interpretability of nonlinear dimensionality reduction.
FeatMAP preserves the source features in low-dimensional embedding space, which facilitates explaining the embedding by high-dimensional features.
Specifically, we embed the tangent space to locally construct a frame showing source features, thus explaining the reduced-dimension results by feature importance.
FeatMAP also maintains the anisotropic density in two-dimensional plot, which further enhances the interpretability of visualization.
The experiments on digit classification, object detection and MNIST adversarial examples demonstrate that featMAP produces interpretable dimensionality reduction results, which boost explaining the classification and feature detection.
Our method to preserve features by tangent space embedding provides a plug-in module for manifold learning.
In future work, we expect to extend our feature preserving paradigm by tangent space embedding to other nonlinear dimensionality reduction methods to enable the interpretability.
Another future work is to apply featMAP to real-world images and biological gene expression data to strengthen interpreting the classification and feature detection.

\bibliographystyle{unsrt}  
\bibliography{references}

\begin{thebibliography}{10}

\bibitem{tenenbaum2000global}
Joshua~B Tenenbaum, Vin~de Silva, and John~C Langford.
\newblock A global geometric framework for nonlinear dimensionality reduction.
\newblock {\em science}, 290(5500):2319--2323, 2000.

\bibitem{roweis2000nonlinear}
Sam~T Roweis and Lawrence~K Saul.
\newblock Nonlinear dimensionality reduction by locally linear embedding.
\newblock {\em science}, 290(5500):2323--2326, 2000.

\bibitem{zhang2006mlle}
Zhenyue Zhang and Jing Wang.
\newblock Mlle: Modified locally linear embedding using multiple weights.
\newblock {\em Advances in neural information processing systems}, 19, 2006.

\bibitem{donoho2003hessian}
David~L Donoho and Carrie Grimes.
\newblock Hessian eigenmaps: Locally linear embedding techniques for
  high-dimensional data.
\newblock {\em Proceedings of the National Academy of Sciences},
  100(10):5591--5596, 2003.

\bibitem{belkin2003laplacian}
Mikhail Belkin and Partha Niyogi.
\newblock Laplacian eigenmaps for dimensionality reduction and data
  representation.
\newblock {\em Neural computation}, 15(6):1373--1396, 2003.

\bibitem{zhang2004principal}
Zhenyue Zhang and Hongyuan Zha.
\newblock Principal manifolds and nonlinear dimensionality reduction via
  tangent space alignment.
\newblock {\em SIAM journal on scientific computing}, 26(1):313--338, 2004.

\bibitem{van2008visualizing}
Laurens Van~der Maaten and Geoffrey Hinton.
\newblock Visualizing data using t-sne.
\newblock {\em Journal of machine learning research}, 9(11), 2008.

\bibitem{mcinnes2018umap}
Leland McInnes, John Healy, and James Melville.
\newblock Umap: Uniform manifold approximation and projection for dimension
  reduction.
\newblock {\em arXiv preprint arXiv:1802.03426}, 2018.

\bibitem{levina2004maximum}
Elizaveta Levina and Peter Bickel.
\newblock Maximum likelihood estimation of intrinsic dimension.
\newblock {\em Advances in neural information processing systems}, 17, 2004.

\bibitem{pope2021intrinsic}
Phillip Pope, Chen Zhu, Ahmed Abdelkader, Micah Goldblum, and Tom Goldstein.
\newblock The intrinsic dimension of images and its impact on learning.
\newblock {\em arXiv preprint arXiv:2104.08894}, 2021.

\bibitem{wright2022high}
John Wright and Yi~Ma.
\newblock {\em High-dimensional data analysis with low-dimensional models:
  Principles, computation, and applications}.
\newblock Cambridge University Press, 2022.

\bibitem{singer2012vector}
Amit Singer and H-T Wu.
\newblock Vector diffusion maps and the connection laplacian.
\newblock {\em Communications on pure and applied mathematics},
  65(8):1067--1144, 2012.

\bibitem{lim2021tangent}
Uzu Lim, Vidit Nanda, and Harald Oberhauser.
\newblock Tangent space and dimension estimation with the wasserstein distance.
\newblock {\em arXiv preprint arXiv:2110.06357}, 2021.

\bibitem{preparata2012computational}
Franco~P Preparata and Michael~I Shamos.
\newblock {\em Computational geometry: an introduction}.
\newblock Springer Science \& Business Media, 2012.

\bibitem{dong2011efficient}
Wei Dong, Charikar Moses, and Kai Li.
\newblock Efficient k-nearest neighbor graph construction for generic
  similarity measures.
\newblock In {\em Proceedings of the 20th international conference on World
  wide web}, pages 577--586, 2011.

\bibitem{liu2016visualizing}
Shusen Liu, Dan Maljovec, Bei Wang, Peer-Timo Bremer, and Valerio Pascucci.
\newblock Visualizing high-dimensional data: Advances in the past decade.
\newblock {\em IEEE transactions on visualization and computer graphics},
  23(3):1249--1268, 2016.

\bibitem{vellido2012making}
Alfredo Vellido, Jos{\'e}~David Mart{\'\i}n-Guerrero, and Paulo~JG Lisboa.
\newblock Making machine learning models interpretable.
\newblock In {\em ESANN}, volume~12, pages 163--172. Bruges, Belgium, 2012.

\bibitem{frenay2016information}
Beno{\^\i}t Fr{\'e}nay and Bruno Dumas.
\newblock Information visualisation and machine learning: Characteristics,
  convergence and perspective.
\newblock In {\em 24th European Symposium on Artificial Neural Networks,
  Computational Intelligence and Machine Learning, ESANN 2016}, pages 623--628.
  i6doc. com publication, 2016.

\bibitem{dumas2018interaction}
Bruno Dumas, Beno{\^\i}t Fr{\'e}nay, and John Lee.
\newblock Interaction and user integration in machine learning for information
  visualisation.
\newblock In {\em 26th European Symposium on Artificial Neural Networks,
  Computational Intelligence and Machine Learning (ESANN 2018)}, pages 97--104.
  i6doc. com. publ., 2018.

\bibitem{gabriel1971biplot}
Karl~Ruben Gabriel.
\newblock The biplot graphic display of matrices with application to principal
  component analysis.
\newblock {\em Biometrika}, 58(3):453--467, 1971.

\bibitem{bibal2019measuring}
Adrien Bibal and Beno{\i}t Fr{\'e}nay.
\newblock Measuring quality and interpretability of dimensionality reduction
  visualizations.
\newblock In {\em Safe Machine Learning Workshop at ICLR}, 2019.

\bibitem{bibal2018finding}
Adrien Bibal, Rebecca Marion, and Beno{\^\i}t Fr{\'e}nay.
\newblock Finding the most interpretable mds rotation for sparse linear models
  based on external features.
\newblock In {\em ESANN}, 2018.

\bibitem{marion2019bir}
Rebecca Marion, Adrien Bibal, and Beno{\^\i}t Fr{\'e}nay.
\newblock Bir: A method for selecting the best interpretable multidimensional
  scaling rotation using external variables.
\newblock {\em Neurocomputing}, 342:83--96, 2019.

\bibitem{sips2009selecting}
Mike Sips, Boris Neubert, John~P Lewis, and Pat Hanrahan.
\newblock Selecting good views of high-dimensional data using class
  consistency.
\newblock In {\em Computer Graphics Forum}, volume~28, pages 831--838. Wiley
  Online Library, 2009.

\bibitem{wu2019solving}
Chieh Wu, Jared Miller, Yale Chang, Mario Sznaier, and Jennifer Dy.
\newblock Solving interpretable kernel dimension reduction.
\newblock {\em arXiv preprint arXiv:1909.03093}, 2019.

\bibitem{bibal2020explaining}
Adrien Bibal, Viet~Minh Vu, G{\'e}raldin Nanfack, and Beno{\^\i}t Fr{\'e}nay.
\newblock Explaining t-sne embeddings locally by adapting lime.
\newblock In {\em ESANN}, pages 393--398, 2020.

\bibitem{bardos2022local}
Avraam Bardos, Ioannis Mollas, Nick Bassiliades, and Grigorios Tsoumakas.
\newblock Local explanation of dimensionality reduction.
\newblock {\em arXiv preprint arXiv:2204.14012}, 2022.

\bibitem{borg2005modern}
Ingwer Borg and Patrick~JF Groenen.
\newblock {\em Modern multidimensional scaling: Theory and applications}.
\newblock Springer Science \& Business Media, 2005.

\bibitem{tang2016visualizing}
Jian Tang, Jingzhou Liu, Ming Zhang, and Qiaozhu Mei.
\newblock Visualizing large-scale and high-dimensional data.
\newblock In {\em Proceedings of the 25th international conference on world
  wide web}, pages 287--297, 2016.

\bibitem{hinton2002stochastic}
Geoffrey~E Hinton and Sam Roweis.
\newblock Stochastic neighbor embedding.
\newblock {\em Advances in neural information processing systems}, 15, 2002.

\bibitem{bohm2022attraction}
Jan~Niklas B{\"o}hm, Philipp Berens, and Dmitry Kobak.
\newblock Attraction-repulsion spectrum in neighbor embeddings.
\newblock {\em Journal of Machine Learning Research}, 23(95):1--32, 2022.

\bibitem{damrich2022contrastive}
Sebastian Damrich, Jan~Niklas B{\"o}hm, Fred~A Hamprecht, and Dmitry Kobak.
\newblock Contrastive learning unifies $ t $-sne and umap.
\newblock {\em arXiv preprint arXiv:2206.01816}, 2022.

\bibitem{amid2019trimap}
Ehsan Amid and Manfred~K Warmuth.
\newblock Trimap: Large-scale dimensionality reduction using triplets.
\newblock {\em arXiv preprint arXiv:1910.00204}, 2019.

\bibitem{wang2021understanding}
Yingfan Wang, Haiyang Huang, Cynthia Rudin, and Yaron Shaposhnik.
\newblock Understanding how dimension reduction tools work: An empirical
  approach to deciphering t-sne, umap, trimap, and pacmap for data
  visualization.
\newblock {\em J. Mach. Learn. Res.}, 22(201):1--73, 2021.

\bibitem{narayan2021assessing}
Ashwin Narayan, Bonnie Berger, and Hyunghoon Cho.
\newblock Assessing single-cell transcriptomic variability through
  density-preserving data visualization.
\newblock {\em Nature Biotechnology}, 39(6):765--774, 2021.

\bibitem{sarfraz2022hierarchical}
Saquib Sarfraz, Marios Koulakis, Constantin Seibold, and Rainer Stiefelhagen.
\newblock Hierarchical nearest neighbor graph embedding for efficient
  dimensionality reduction.
\newblock In {\em Proceedings of the IEEE/CVF Conference on Computer Vision and
  Pattern Recognition}, pages 336--345, 2022.

\bibitem{zu2022spacemap}
Xinrui Zu and Qian Tao.
\newblock Spacemap: Visualizing high-dimensional data by space expansion.
\newblock In {\em International Conference on Machine Learning}, pages
  27707--27723. PMLR, 2022.

\bibitem{guo2022co}
Yunhui Guo, Haoran Guo, and Stella~X Yu.
\newblock Co-sne: Dimensionality reduction and visualization for hyperbolic
  data.
\newblock In {\em Proceedings of the IEEE/CVF Conference on Computer Vision and
  Pattern Recognition}, pages 21--30, 2022.

\bibitem{jolliffe2016principal}
Ian~T Jolliffe and Jorge Cadima.
\newblock Principal component analysis: a review and recent developments.
\newblock {\em Philosophical Transactions of the Royal Society A: Mathematical,
  Physical and Engineering Sciences}, 374(2065):20150202, 2016.

\bibitem{coifman2005geometric}
Ronald~R Coifman, Stephane Lafon, Ann~B Lee, Mauro Maggioni, Boaz Nadler,
  Frederick Warner, and Steven~W Zucker.
\newblock Geometric diffusions as a tool for harmonic analysis and structure
  definition of data: Diffusion maps.
\newblock {\em Proceedings of the national academy of sciences},
  102(21):7426--7431, 2005.

\bibitem{deng2020low}
Tingquan Deng, Dongsheng Ye, Rong Ma, Hamido Fujita, and Lvnan Xiong.
\newblock Low-rank local tangent space embedding for subspace clustering.
\newblock {\em Information Sciences}, 508:1--21, 2020.

\bibitem{lecun1998mnist}
Yann LeCun.
\newblock The mnist database of handwritten digits.
\newblock {\em http://yann. lecun. com/exdb/mnist/}, 1998.

\bibitem{xiao2017fashion}
Han Xiao, Kashif Rasul, and Roland Vollgraf.
\newblock Fashion-mnist: a novel image dataset for benchmarking machine
  learning algorithms.
\newblock {\em arXiv preprint arXiv:1708.07747}, 2017.

\bibitem{nene1996columbia}
Sameer~A Nene, Shree~K Nayar, Hiroshi Murase, et~al.
\newblock Columbia object image library (coil-20).
\newblock 1996.

\bibitem{krizhevsky2009learning}
Alex Krizhevsky, Geoffrey Hinton, et~al.
\newblock Learning multiple layers of features from tiny images.
\newblock 2009.

\bibitem{liu2021time}
Can Liu, Andrew~J Martins, William~W Lau, Nicholas Rachmaninoff, Jinguo Chen,
  Luisa Imberti, Darius Mostaghimi, Danielle~L Fink, Peter~D Burbelo, Kerry
  Dobbs, et~al.
\newblock Time-resolved systems immunology reveals a late juncture linked to
  fatal covid-19.
\newblock {\em Cell}, 184(7):1836--1857, 2021.

\bibitem{goodfellow2014explaining}
Ian~J Goodfellow, Jonathon Shlens, and Christian Szegedy.
\newblock Explaining and harnessing adversarial examples.
\newblock {\em arXiv preprint arXiv:1412.6572}, 2014.

\bibitem{van2014accelerating}
Laurens Van Der~Maaten.
\newblock Accelerating t-sne using tree-based algorithms.
\newblock {\em The Journal of Machine Learning Research}, 15(1):3221--3245,
  2014.

\bibitem{espadoto2019toward}
Mateus Espadoto, Rafael~M Martins, Andreas Kerren, Nina~ST Hirata, and
  Alexandru~C Telea.
\newblock Toward a quantitative survey of dimension reduction techniques.
\newblock {\em IEEE transactions on visualization and computer graphics},
  27(3):2153--2173, 2019.

\bibitem{venna2006local}
Jarkko Venna and Samuel Kaski.
\newblock Local multidimensional scaling.
\newblock {\em Neural Networks}, 19(6-7):889--899, 2006.

\bibitem{venna2006visualizing}
Jarkko Venna and Samuel Kaski.
\newblock Visualizing gene interaction graphs with local multidimensional
  scaling.
\newblock In {\em ESANN}, volume~6, pages 557--562, 2006.

\bibitem{joia2011local}
Paulo Joia, Danilo Coimbra, Jose~A Cuminato, Fernando~V Paulovich, and Luis~G
  Nonato.
\newblock Local affine multidimensional projection.
\newblock {\em IEEE Transactions on Visualization and Computer Graphics},
  17(12):2563--2571, 2011.

\bibitem{lecun1998gradient}
Yann LeCun, L{\'e}on Bottou, Yoshua Bengio, and Patrick Haffner.
\newblock Gradient-based learning applied to document recognition.
\newblock {\em Proceedings of the IEEE}, 86(11):2278--2324, 1998.

\end{thebibliography}

\section{Supplementary}
\subsection{Stochastic gradient descent (SGD) for anisotropic projection}
We use anisotropic projection to embed the data points into low-dimensional space.
To apply SGD to optimizing the embedding coordinates, we calculate the derivative of the loss function
\begin{equation}
\label{eq:loss_f}
\mathcal{L} = \mathit{CE}(P||Q) - \lambda\mathit{Corr}(r^o, r^e).
\end{equation}

The core of the anisotropic projection lies in optimizing the Pearson correlation between the local radius in original and embedding space. We first calculate the gradient of this correlation regarding the embedding coordinates for optimization.
We rewrite the correlation as follows:
\begin{equation}
\label{eq:corr}
\mathit{Corr}(r^o, r^e) = \sum_{l=1}^{d'} \frac{Cov(r^o_l, r^e_l)}{(\mathit{Var}(r_l^o)\mathit{Var}(r_l^e))^{1/2}}
\end{equation}
where $r_l^o = \{r_{il}^o \}_{i=1}^m = \{\mathit{log} R_{il}^o \}_{i=1}^m$ and 
$r_l^e = \{r_{il}^e \}_{i=1}^m = \{\mathit{log} R_{il}^e \}_{i=1}^m$ in the $l$-th principal direction of original and embedding spaces respectively.
% Let $R_{il}^o$ and $R_{il}^e, i=1,...,m$ denote the local radius in the $l$-th principal direction of original and embedding spaces respectively. 
% Let $r_{il}^e = \mathit{log} R_{il}^e$. We center the original density and set $r_{il}^o = \mathit{log}R_{il}^o - m^{-1}\sum_{i=1}^m \mathit{log}R_{il}^o$.
We set $z_{ij} = y_j - y_i$ and rewrite $r_{il}^e$ as
\begin{equation}
\label{eq:r_emb}
\begin{split}
     r_{il}^e = \mathit{log} R_{il}^e &= \mathit{log} \frac{1}{\sum_j Q_{ij}}\sum_j Q_{ij}\|(y_j-y_i)v_{il}^e \|^2 \\
    &= \mathit{log} \frac{1}{\sum_j Q_{ij}}\sum_j Q_{ij}\|z_{ij}v_{il}^e \|^2,
\end{split}
\end{equation}
and
\begin{equation}
    Q_{ij} = \frac{1}{1+a(y_j - y_i)^{2b}} = [1+a(z_{ij}^T z_{ij})^b]^{-1}
\end{equation}
Let
\begin{equation}
    \rho_{e,o}^{(l)} = \frac{Cov(r^o_l, r^e_l)}{(\mathit{Var}(r_l^o)\mathit{Var}(r_l^e))^{1/2}},
\end{equation}
be the correlation in the $l$-th direction.
The derivative of $\rho_{e,o}^{(l)}$ with respect to $z_{ij}$ is
\begin{equation}
\label{eq:grad}
% \begin{split}
    \frac{\partial \rho_{e,o}^{(l)} }{\partial z_{ij}} = \mathit{Var}(r_l^o)^{-1/2} 
[\frac{\partial \mathit{Cov}(r_l^o, r_l^e)}{\partial z_{ij}} \mathit{Var}(r_l^e)^{-1/2} - 
\frac{1}{2} \mathit{Cov}(r_l^o, r_l^e) \mathit{Var}(r_l^e)^{-3/2} 
\frac{\partial \mathit{Var}(r_l^e)}{\partial z_{ij}} ]
% \end{split}
\end{equation}
Therefore, the gradient of the correlation in~\Cref{eq:corr_1} in the $l$-th principal direction regarding the embedding coordinates $y_i$ is 
\begin{equation}
\label{eq:grad_corr}
\Delta_{y_i} \mathit{Corr}^{(l)}(r^o, r^e) = \sum_{j\neq i} \frac{\partial \rho_{e,o}^{(l)} }{\partial z_{ij}}  \frac{\partial z_{ij}}{\partial y_i}
\end{equation}
We further compute the derivative of the variance and covariance in~\Cref{eq:grad}.
To simplify the notation, we set $\mu_l^e = \mathbb{E}[r_l^e]$ and center the original local radius as $r_{il}^o:= \mathit{log} R_{il}^o - m^{-1} \sum_{i=1}^m \mathit{log} R_{il}^o$.
The gradient in~\Cref{eq:grad} ($l$ omitted) is calculated as
\begin{equation}
\label{eq:corr_grad}
\begin{split}
\frac{\partial \rho_{e,o}^{(l)} }{\partial z_{ij}} &= \mathit{Var}(r^o)^{-\frac{1}{2}} 
\{\frac{1}{m-1} (r_i^o \frac{\partial r_i^e}{\partial z_{ij}} + r_j^o \frac{\partial r_j^e}{\partial z_{ij}} ) \mathit{Var}(r_l^e)^{-\frac{1}{2}}\\
&- \frac{1}{m-1} \mathit{Cov}(r^o, r^e) \mathit{Var}(r^e)^{-\frac{3}{2}}[(r_i^e - \mu^e) \frac{\partial r_i^e}{\partial z_{ij}} \\
&+(r_j^e - \mu^e) \frac{\partial r_j^e}{\partial z_{ij}} ]  \},
\end{split}
\end{equation}
and 
\begin{equation}
% \begin{split}
\frac{\partial r_i^e}{\partial z_{ij} } = \frac{\Tilde{Q}_{ij}^2 W_i } {R_i^e}[(1+a(z_{ij}^T z_{ij})^b - 2ab(z_{ij}^T z_{ij})^{b-1} z_{ij} z_{ij}^T )\\
2 v_i v_i^T z_{ij}] + \Tilde{Q}_{ij}^2 W_i * 2ab(z_{ij}^T z_{ij})^{b-1} z_{ij}
% \end{split}
\end{equation}
where $W_i = \sum_{s=1}(1+a (z_{is}^T z_{is})^b)^{-1}$ and 
$\Tilde{Q}_{ij} = W_i^{-1} (1+a (z_{ij}^T z_{ij})^b)^{-1}$.

Next, we compute the gradient of the cross-entropy $\mathit{CE}(P||Q)$ (similar to UMAP~\cite{mcinnes2018umap}).
The cross-entropy loss function is 
\begin{equation}
    \mathcal{L}_{ce} = \mathit{CE}(P||Q) = -\sum_{ij} P_{ij} \mathit{log} Q_{ij} + (1-P_{ij}) \mathit{log}(1-Q_{ij})
\end{equation}
and its gradient with respect to $z_{ij}$ is
\begin{equation}
\label{eq:ce_grad}
    \frac{\partial \mathcal{L}_{ce}}{\partial z_{ij}} = -[P_{ij} \frac{\partial \mathit{log}Q_{ij}}{\partial z_{ij}} + 
    (1-P_{ij}) \frac{\partial \mathit{log}(1-Q_{ij}) }{\partial z_{ij}}]
\end{equation}
where $\frac{\partial \mathit{log}Q_{ij}}{\partial z_{ij}} = 
\frac{-2ab(z_{ij}^T z_{ij})^{b-1} z_{ij}}{1+a(z_{ij}^T z_{ij})^b}$
and $\frac{\partial \mathit{log}(1-Q_{ij}) }{\partial z_{ij}} =
\frac{1}{z_{ij}^Tz_{ij}} \frac{2b z_{ij}}{1+ (z_{ij}^T z_{ij})^b}$.

The attractive term $P_{ij} \mathit{log} Q_{ij}$ is optimized by randomly drawing an edge $(i,j)$ by distribution $P$, which means that the edge $(i,j)$ is selected with probability $\frac{P_{ij}}{\sum_{i\neq j} P_{ij} }$, followed by calculating the gradient $\frac{\partial \mathit{log}Q_{ij}}{\partial z_{ij}}$.
The repulsive term is estimated by uniformly at random choosing a set of points $S$ adjacent to the given point $x_i$ and computing $\frac{1}{|S|} \sum_{l\in S} \frac{\partial \mathit{log}(1-Q_{il})}{\partial z_{il}}$.

Combine the gradient of cross-entropy in~\Cref{eq:ce_grad} and local radius correlation in~\Cref{eq:corr_grad}. The  gradient for an edge $(i,j)$  at each iteration of the SGD  in the $l$-th principal direction is
\begin{equation}
% \begin{split}
    \Delta_{y_i} \mathcal{L} |_{(i,j)} = 
    -[\frac{\partial \mathit{log}Q_{ij}}{\partial z_{ij}} -
     \frac{1}{|S|} \sum_{l\in S} \frac{\partial \mathit{log}(1-Q_{il})}{\partial z_{il}} \\
    +\lambda \frac{Z}{m P_{ij}} \frac{\partial \rho_{e,o}^{(l)} }{\partial z_{ij}} ]  \frac{\partial z_{ij}}{\partial y_i},
% \end{split}
\end{equation}
where $Z = \sum_{(i,j)} P_{ij}$ and $\frac{m P_{ij}}{Z}$ is the normalization factor considering the edge is chosen with probability $\frac{ P_{ij}}{Z}$.

In addition, we compute the gradient in~\Cref{eq:corr_grad} at the start of each epoch. 
We achieve this by calculating $W_i = \sum_{s=1}Q_{is}$, the local radius $r_{il}^e$, the variance and covariance terms in the beginning of each epoch, regarding them as fixed for all the edges to be updated during that epoch.

\subsection{Stochastic Gradient Descent (SGD) for tangent space embedding}
The key of tangent space embedding is to preserve the alignment between tangent spaces.
We depict the alignment by the general angle between two tangent spaces, which induces the cosine distance of tangent spaces. 
We further model the probability distribution of cosine distance in original and embedding space respectively, and minimize the difference of distribution by KL divergence to maintain the alignment in embedding space.

In practice, we borrow the framework of t-SNE or UMAP to compute the embedding tangent space. Specifically, we flat the tensor of tangent space in original space and feed it to t-SNE as input. We set the parameter \textit{distance} as \textit{cosine}. The output is normalized to get the embedding tangent space. 

\subsection{FeatMAP interpreting  adversarial examples}
\begin{figure*}[t]
  \centering
%   \fbox{\rule{0pt}{2in} \rule{0.9\linewidth}{0pt}}
  \includegraphics[width=1.0\linewidth]{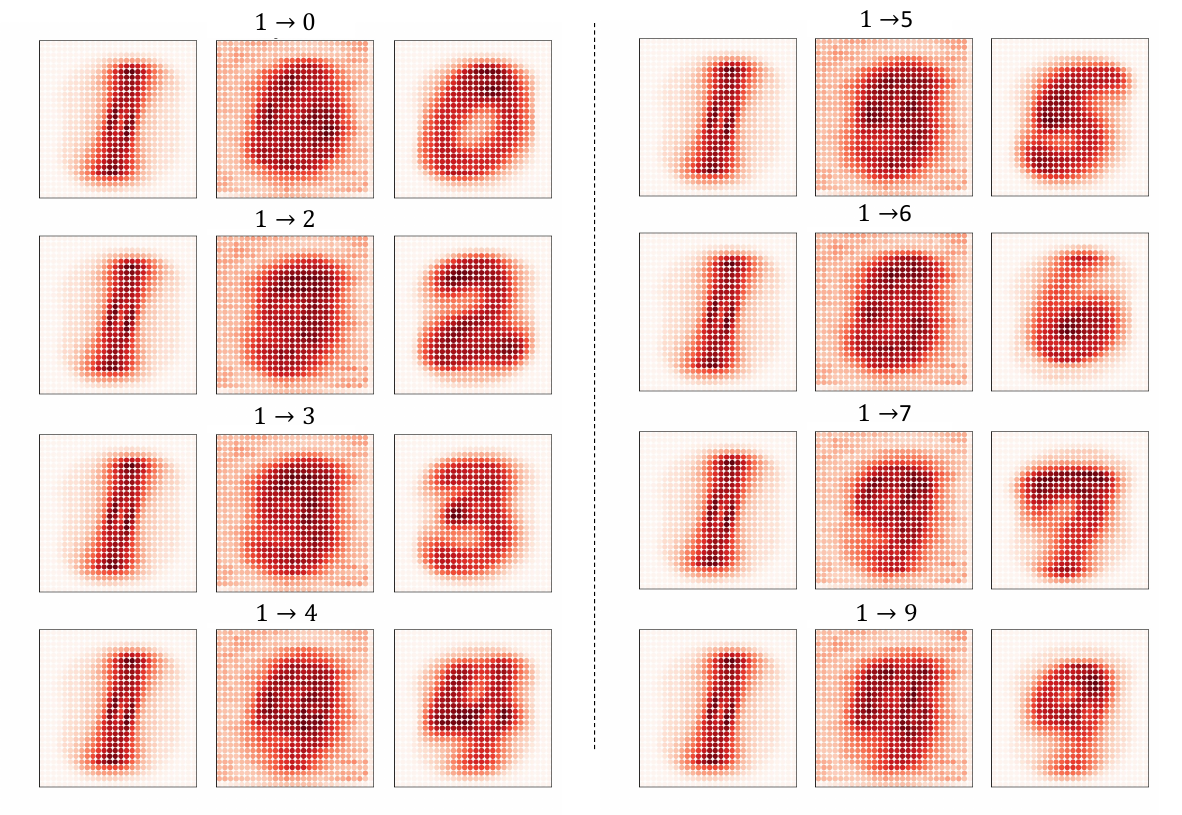}

  \caption{Average feature importance of the adversarial examples and the corresponding original data.
  For the original digits $1$, we list all the corresponding adversarial examples with average feature importance in the middle. 
%   The adversarial examples with average feature importance are in the middle. 
  The left is the original digit $1$ and the right is the original digit with the label same as the corresponding adversarial label. }
  \label{fig:adv_other}. 
\end{figure*}

We illustrate how the digits $1$ are misclassified to the other labels in~\Cref{fig:adv_other}. The saliency map illustrates the average feature importance.
The feature importance pattern in the middle is more similar to the right than the left, 
indicating that the adversarial attack FGSM alters the feature importance and fools the LeNet classifier.

\subsection{FeatMAP preserving original density}
\begin{figure*}[t]
  \centering
%   \fbox{\rule{0pt}{2in} \rule{0.9\linewidth}{0pt}}
  \includegraphics[width=1.0\linewidth]{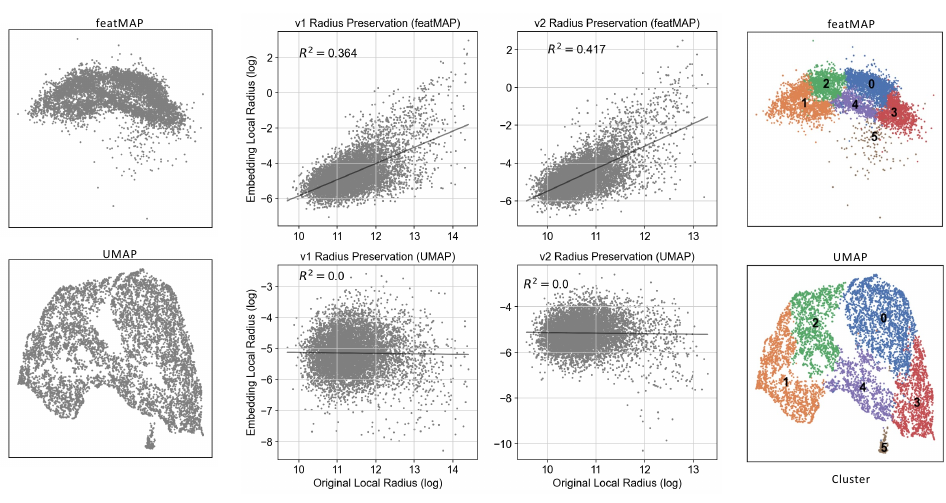}

  \caption{FeatMAP preserving the original density.
  FeatMAP and UMAP are applied to digit $1$ group of MNIST (left). 
  The middle is the scatter plot of embedding local radius against original local radius, along with the straight line of linear regression showing the correlation of local radius between embedding and original space. 
  The right illustrates the clusters on both featMAP and UMAP.
  }
  \label{fig:den_1}
\end{figure*}

\begin{figure*}[t]
  \centering
%   \fbox{\rule{0pt}{2in} \rule{0.9\linewidth}{0pt}}
  \includegraphics[width=1.0\linewidth]{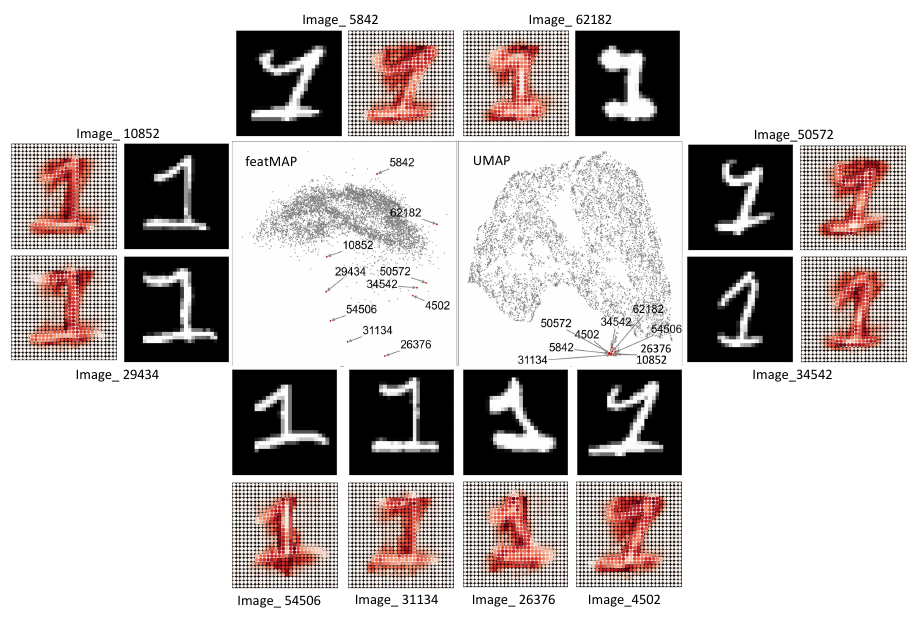}

  \caption{Handwritten digit images from the sparse subgroup of featMAP.
  $10$ data points are randomly selected from the subgroup $5$ in~\Cref{fig:den_1} with the corresponding digit images.
  }
  \label{fig:sparse}
\end{figure*}

FeatMAP presents larger local radius correlation than UMAP in~\Cref{fig:den_1}.
% indicating that featMAP performs better in local density preservation.
% We cluster the dataset by~\cite{traag2019louvain}.
% in Figure xx (Supplementary). 
We cluster the data (digit $1$ in MNIST) on the right of~\Cref{fig:den_1} and find that the subgroup $5$ (in brown colour) is significantly more sparse in featMAP than UMAP.
% We claim that featMAP correctly reveals this sparse pattern because featMAP shows positive correlation of local radius between embedding and original space.
% We cluster the dataset and find that featMAP correctly shows the positive correlation of local radius between embedding and original space for the subgroup (in brown colour) in ~\Cref{fig:density}.
% The plot of this subgroup by featMAP presents more sparse pattern than UMAP.
% the subgroup (in brown colour) is significantly more sparse in featMAP than UMAP in~\Cref{fig:density}.
% and local radius correlation in featMAP is larger than UMAP in~\Cref{fig:density}. 
We further demonstrate the digit images from this subgroup in~\Cref{fig:sparse}, which exhibit diverse handwritten patterns and agree with the sparse pattern in featMAP.

\subsection{Hyperparameter tuning}
We demonstrate the tuning of the hyperparameter $\lambda$ in the loss function of featMAP in~\Cref{fig:lambda}.
The parameter $\lambda=0$ corresponds to UMAP.
With the increasing of $\lambda$, featMAP performs better in preserving the local anisotropic density with the local radius correlation getting larger, while the clusters in the two-dimensional plot are not clearly separable when $\lambda$ is large. 
Consider the trade-off between the local density preservation and clusters visualization, we set the parameter $\lambda$ as $0.5$.

\afterpage{
\begin{figure*}[t]
  \centering
%   \fbox{\rule{0pt}{2in} \rule{0.9\linewidth}{0pt}}
  \includegraphics[width=1\linewidth]{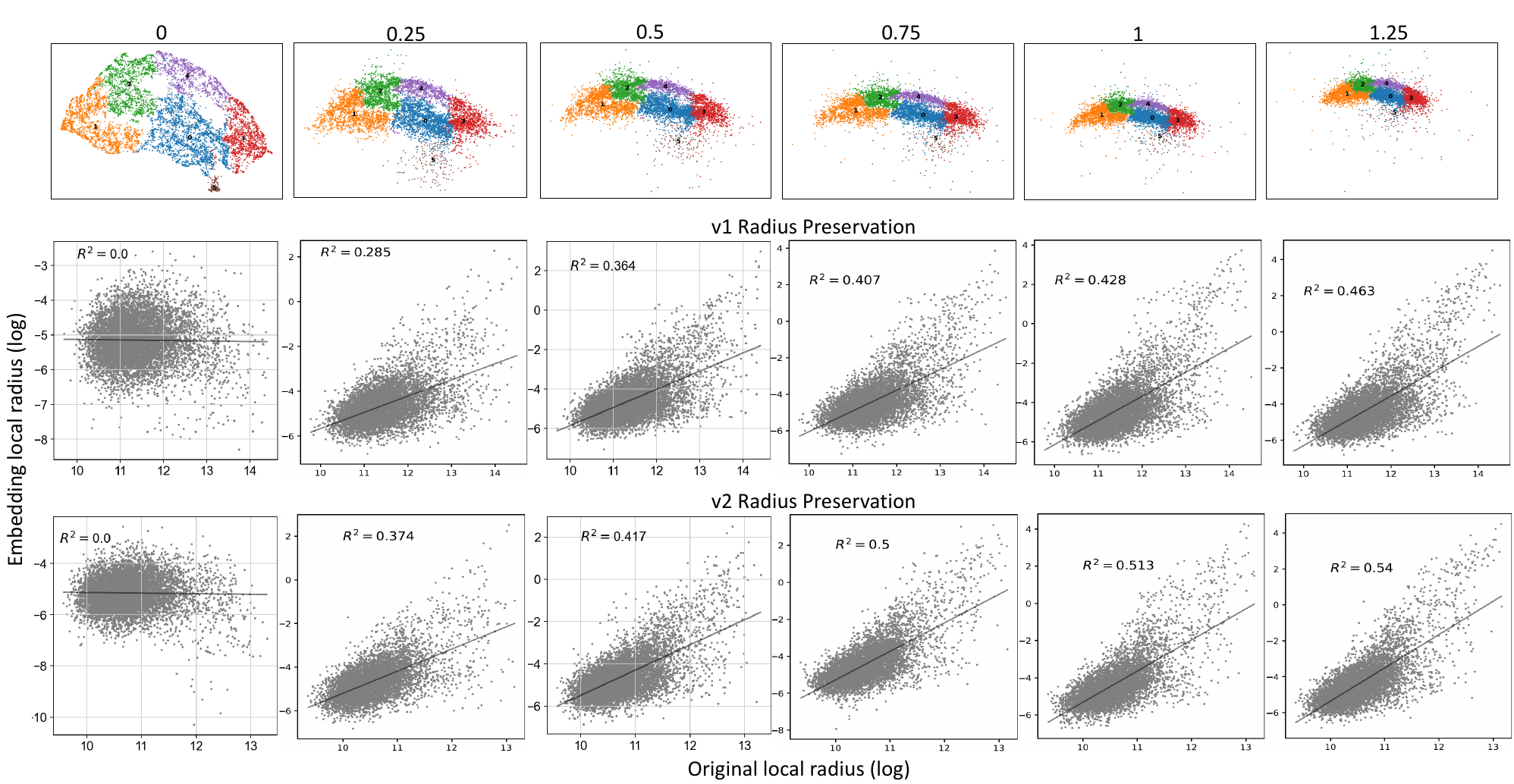}

  \caption{The tuning of hyperparameter $\lambda$.
  The hyperparameter $\lambda$ in featMAP is set from $0$ to $1.25$ with two-dimensional plot and local radius correlation.
  }
  \label{fig:lambda}
\end{figure*}
\clearpage
}
\subsection{Comparing featMAP with other state-of-the-art}
We apply the state-of-the-art nonlinear dimensionality reduction methods to the benchmark datasets to visualize the data in~\Cref{fig:overall_plot}.
FeatMAP illustrates the density preservation as well as densMAP~\cite{narayan2021assessing}.

\afterpage{
\begin{figure*}[t]
  \centering
%   \fbox{\rule{0pt}{2in} \rule{0.9\linewidth}{0pt}}
  \includegraphics[width=1.0\linewidth]{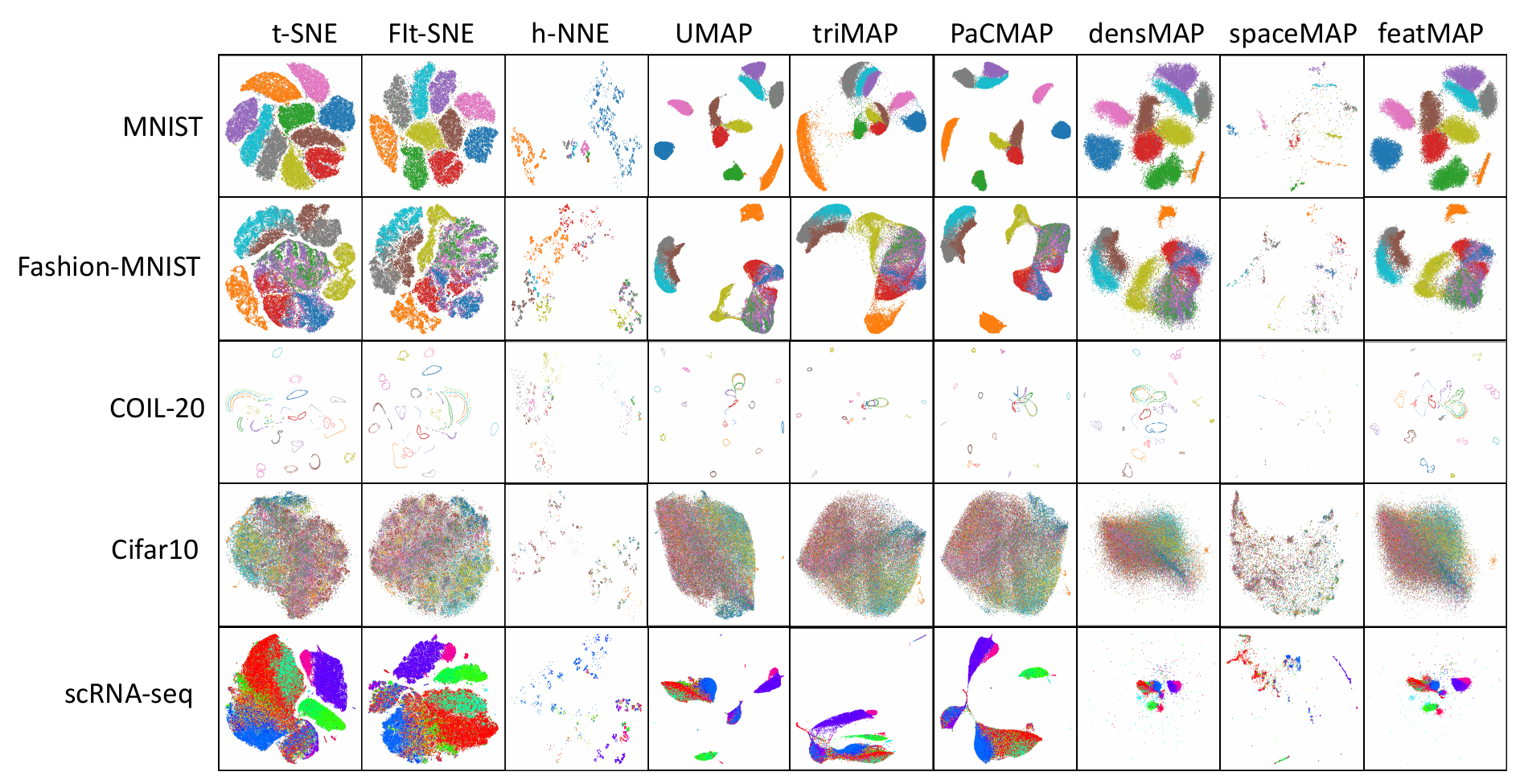}

  \caption{Visualization of multiple datasets by the state-of-the-art nonlinear dimensionality reduction methods.
  }
  \label{fig:overall_plot}
\end{figure*}
\clearpage
}

\end{document}